 \documentclass[pmlr,twocolumn,10pt]{jmlr} 

\usepackage{mathtools}

\usepackage{graphicx}
\usepackage{glossaries}

\DeclarePairedDelimiterX{\kldivx}[2]{(}{)}{%
  #1\;\delimsize\|\;#2%
}
\newcommand{\kldiv}{\mathrm{KL}\kldivx}

\setacronymstyle{long-short}
\newacronym{ehr}{EHR}{electronic health record}
\newacronym{rnn}{RNN}{recurrent neural network}
\newacronym{lps-co}{LPS-CO}{longitudinal patient stratification by clinical outcomes}
\newacronym{gru}{GRU}{gated recurrent unit}
\newacronym{vae}{VAE}{variational autoencoder}
\newacronym{ari}{ARI}{adjusted Rand index}
\newacronym{km}{KM}{Kaplan-Meier}
\newacronym{pca}{PCA}{principal component analysis}
\newacronym{dec}{DEC}{deep embedded clustering}
\newacronym{rsf}{RSF}{random survival forest}
\newacronym{kl}{KL}{Kullback-Leibler}

\glsdisablehyper

\usepackage{booktabs}
\usepackage[load-configurations=version-1]{siunitx} 

\newcommand{\equal}[1]{{\hypersetup{linkcolor=black}\thanks{#1}}}

\theorembodyfont{\upshape}
\theoremheaderfont{\scshape}
\theorempostheader{:}
\theoremsep{\newline}

\jmlrvolume{LEAVE UNSET}
\jmlryear{2021}
\jmlrsubmitted{LEAVE UNSET}
\jmlrpublished{LEAVE UNSET}
\jmlrworkshop{Machine Learning for Health (ML4H) 2021} 

\title[Longitudinal patient stratification with clinical outcomes]{Longitudinal patient stratification of electronic health records with flexible adjustment for clinical outcomes}

\author{%
\Name{Oliver Carr}\equal{These authors contributed equally}  \Email{oliver.carr@sensynehealth.com}\\
\Name{Avelino Javer}\footnotemark[1] \Email{avelino.javer@sensynehealth.com}\\
\Name{Patrick Rockenschaub}\footnotemark[1]  \Email{patrick.rockenschaub@sensynehealth.com}\\
\Name{Owen Parsons} \Email{owen.parsons@sensynehealth.com}\\
\Name{Robert D{\"u}richen}  \Email{robert.durichen@sensynehealth.com}\\
\addr Sensyne Health Plc, Oxford, UK
 }

\begin{document}

\maketitle

\begin{abstract}
The increase in availability of longitudinal \gls{ehr} data is leading to improved understanding of diseases and discovery of novel phenotypes. The majority of clustering algorithms focus only on patient trajectories, yet patients with similar trajectories may have different outcomes. Finding subgroups of patients with different trajectories and outcomes can guide future drug development and improve recruitment to clinical trials. We develop a recurrent neural network autoencoder to cluster \gls{ehr} data using reconstruction, outcome, and clustering losses which can be weighted to find different types of patient clusters. We show our model is able to discover known clusters from both data biases and outcome differences, outperforming baseline models. We demonstrate the model performance on $29,229$ diabetes patients, showing it finds clusters of patients with both different trajectories and different outcomes which can be utilized to aid clinical decision making.
\end{abstract}

\section{Introduction}

Chronic diseases like diabetes or heart failure may progress very differently across patients  \citep{Spratt2017, Lewis2017} but the reasons for differential progression are not yet well understood. Between patient differences have been linked --- among other factors --- to the underlying pathology or to differential response to treatment \citep{Sarria-Santamera2020, Hedman2020}.

Over the last decade, the spread of \glspl{ehr} has enabled the collection of unprecedented longitudinal patient information \citep{Shickel2018}. Early work using this rich data to investigate heterogeneous disease progression mostly employed \emph{unsupervised clustering} to find patient subgroups in the data that share a similar medical history \citep{Miotto2016, Baytas2017, Madiraju2018, Landi2020}. However, \gls{ehr} data are primarily designed for clinical care and are not usually collected with research in mind. Identified clusters may therefore be driven by spurious associations such as patient drop out, selective recording, modifications of the IT infrastructure, or administrative differences between healthcare providers \citep{DeJong2019, Ehrenstein2019}.  

In an attempt to address these issues, new methods have been developed that focus on relevant patient outcomes to guide the clustering of patient trajectories \citep{Zhang2019, Lee2020, Lee2020a}, e.g., by including occurrence of complications or time to death. In this so-called \emph{predictive clustering}, a low-dimensional latent representations of the data is created that retains only information predictive of future clinical events. Patients are grouped according to their similarity in this latent space. While this approach ensures clusters that differ in the risk of experiencing the outcome, they are unable to distinguish between distinct trajectories that lead to similar risks.

Retaining trajectories, however, can be paramount to clinical interpretation. 
For example, although patients with acute heart failure may have short-term mortality risks that are very similar to patients hospitalised with sepsis, the mechanisms that cause the high risk are quite different and a model should be able to distinguish between them. 
In this work, we therefore propose a novel semi-supervised architecture that combines both approaches --- predictive and unsupervised --- to guide clustering towards outcomes of interest while enforcing similarity on the input scale. 
By doing so, we ensure that patients with very different trajectories are not lumped into a common cluster but remain in separate groups that facilitate clinical interpretation. 
Changing the weights of the unsupervised and predictive optimisation functions, the algorithm can be adjusted to prioritise one or the other.
We refer to this approach as \gls{lps-co}.

We apply our method to right-censored clinical data --- which is ubiquitous in \gls{ehr} data --- and show how it can lead to novel insights. 

Our main contributions can be summarised as follows:
\begin{itemize}
    \item Introduction of a flexible semi-supervised patient stratification approach which identifies clusters of patients which share a similar medical history as well as clinical outcomes through parallel optimisation of an unsupervised and predictive loss function.
    \item Introduction of a Cox proportional hazards loss function to consider right-censored outcomes as predictive targets such as time to death or re-hospitalisation.
\end{itemize}

We validate our proposed method on a synthetic dataset with known clusters as well as on a diabetes cohort extracted from a longitudinal \gls{ehr} dataset consisting of approximately half a million patients. 
Comparisons to other baseline methods indicate how our approach can balance between unsupervised and predictive clustering and discover novel patient clusters.

\section{Related Work}

Initial work in patient phenotyping mostly applied clustering to cross-sectional data. Patient phenotyping using k-means has been used early for example in diabetes \citep{Hammer2003} and heart failure \citep{Ather2009}. Other commonly applied methods include hierarchical clustering \citep{Moore2010, Burgel2010} and self-organising maps \citep{Ather2009}. Recently, these have been partially superseded by methods based on autoencoders, which provide an elegant way to deal with increasingly high-dimensional medical data. Notably, \cite{Xie2015-mw} proposed a \gls{dec} algorithm that uses an autoencoder with a self-supervised loss function to jointly learn the low-dimensional representation and cluster assignments. This approach has been used in \cite{Carr2020} and \cite{CastelaForte2021}, among others, and provides the basis for our proposed approach.

With the advent of \glspl{ehr} and increasing availability of longitudinal patient data, unsupervised methods have also been used for phenotyping of sequential medical data. Proposed models include generalisations of classical methods (see for example \cite{Mullin2021}) as well as deep learning-based algorithms to longitudinal data. In the latter case, recurrent autoencoders \citep{Zhang2018, DeJong2019} or convolutions \citep{zhu2016} have been used to embed the time series.

When evaluating the groups identified during clustering via the above methods, patients are often compared based on the risk of experiencing clinically relevant outcomes. For example, \cite{CastelaForte2021} show that among intensive care patients, cluster membership was associated with risk of death. The analysis is entirely post-hoc, however, and differences in risk did not directly influence the earlier cluster assignments. Recent works have aimed to incorporate information of outcomes into the discovery of clusters. \cite{Zhang2019}  used a \gls{rnn} to predict markers of progression in Parkinson's disease and then employed dynamic time warping \citep{Berndt1994} and t-distributed Stochastic Neighbor Embedding (t-SNE) \citep{Van_der_Maaten2008} to cluster patients based on the hidden states of the \gls{rnn}. \cite{Lee2020} proposed an actor-critic approach for temporal predictive clustering (AC-TPC) in which an \gls{rnn}-based encoder/predictor network is trained jointly with the cluster embeddings. This was extended in \cite{Lee2020a} to incorporate time-to-event outcomes via a novel loss function based on a Weibull-shaped parametric hazard. 

In our work, we have adapted a \gls{rnn} autoencoder through the addition of a clustering loss \citep{Xie2015-mw} and an outcome loss \citep{Bello2019} and propose flexible balancing of these losses, thereby allowing researchers to control the degree to which clinical outcomes should drive the clustering. This differs from \cite{Zhang2019, Lee2020, Lee2020a} who focus on outcomes without retaining trajectory information in the clusters.


\section{Methods}

This section describes the methods and model architectures used to obtain patient representations from patient trajectories and the clustering methods applied.

Let $\mathcal{D} = \{ \mathcal{X}, \mathcal{Y} \}^N_{n=1}$ define the patient data, where $\mathcal{X}$ is a set of covariate vectors, $\mathcal{Y}$ is a set of clinical outcomes, and $N$ is the total number of patients included in the data. $\mathcal{D}$ may describe each patient $n$'s observations at a single point of time $\{\textbf{x}^n, y^n\}$ or longitudinally over a period of time $\{ \textbf{x}^n_t, y^n_t \}_{t=1}^{T^n}$. Similarly, $\mathcal{Y}$ may contain continuous outcomes  $y^n \in \mathbb{R}$, binary outcomes $y^n \in \{0, 1\}$, or time-to-event outcomes $y^n = \{ s^n, c^n \}$ with $s^n \in \mathbb{R}^+$ being the patient's follow-up time and $c^n \in \{0, 1\}$ being an indicator of whether the patient experienced the outcome of interest (1) or was censored (0). Going forward and without loss of generality, we omit the time subscript $t$ for simplicity and assume time-to-event data.  

Following \cite{Xie2015-mw}, we aim to cluster $\mathcal{X}$ into $K$ clusters, each of which is represented by a centroid vector $\lambda_k$ for $k = 1,...,K$. In order to deal with the challenges posed by high dimensionality frequently observed in \gls{ehr} data, clustering is not performed in the input space but in a latent embedding space created by a learned, non-linear function $f_{\theta} : \mathcal{X} \rightarrow \mathcal{Z}$.

\begin{figure}
\centering
\includegraphics[width=\columnwidth]{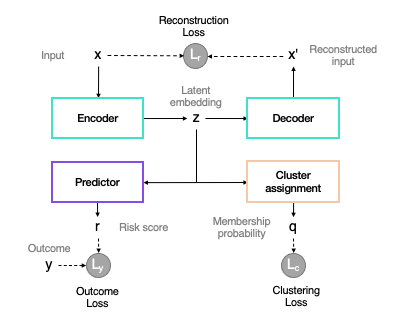}
\caption{Schematic diagram of proposed method consisting of an autoencoder network (encoder and decoder), a predictor layer, and a cluster assignment layer.}
\label{fig:architecture}
\end{figure}

We combined three losses with corresponding weights to give the overall loss function,

\begin{equation}
    L = w_rL_r + w_yL_y + w_cL_c,
\label{eq:loss}
\end{equation}

\noindent
where $L_r$ represents a reconstruction loss, $L_{y}$ an outcome loss, and $L_c$ a clustering loss (Figure \ref{fig:architecture}). $w_*$ represents the weight for each loss. The network architecture and losses are described in detail in the following sections.

\subsection{Patient Embedding}

In our proposed method, we first train an autoencoder network and obtain a fixed-size latent embedding vector $\textbf{z}^n$ for each patient $n$ in $\mathcal{D}$ \citep{Hinton2006}. In it's simplest form, the autoencoder consists of a fully-connected encoder network $f_{\theta}: \mathbb{R}^D \rightarrow \mathbb{R}^{D'}$ and a fully-connected decoder network $g_{\theta'}: \mathbb{R}^{D'} \rightarrow \mathbb{R}^D$, where $D$ is the dimension of the input space and $D'$ is the dimension of the latent space, with $D' < D$. The autoencoder is trained in an unsupervised manner to maximise the information about $\textbf{x}$ retained in $\textbf{z} = f_{\theta}(\textbf{x})$ \citep{Vincent2010}. This is achieved by minimising a reconstruction loss $L_r(\textbf{x}, \textbf{x}')$, where $\textbf{x}' = g_{\theta'}(\textbf{z})$ is reconstructed from the latent embedding. In the case of mixed continuous and binary input, $L_r$ can be defined as 

\begin{equation}
    L_r(\textbf{x}, \textbf{x}') = L_{cont} + w_b * L_{bin} 
    \label{eq:lrr}
\end{equation}
\begin{equation}
    L_{cont} = \frac{1}{N}\sum_{n=1}^{N} (\mathbf{x}_{cont}^n-\mathbf{x'}^n_{cont})^2
\label{eq:lrcon}
\end{equation}
\begin{align}
    L_{bin} =\frac{1}{N}\sum_{n=1}^{N}  \Big\{  & \mathbf{x}^n_{bin}~log(\mathbf{x'}^n_{bin}) \nonumber \\ 
    & - (1-\mathbf{x}^n_{bin}) ~log(1-\mathbf{x'}^n_{bin}) \Big\} 
     \label{eq:lrbin}
\end{align}

\noindent
where Equation \ref{eq:lrcon} is the mean squared error of all continuous inputs $\textbf{x}_{cont}$, Equation \ref{eq:lrbin} is the binary cross entropy of all binary inputs $\textbf{x}_{bin}$, and $w_b$ is a weight to balance the relative contributions of each loss.

\subsubsection{Alternative autoencoders}

The above autoencoder can be extended to sequential patient data $\mathcal{X}_t$ by replacing the fully-connected encoder and decoder networks with a \gls{rnn}. The term \gls{rnn} may in this case represent any recurrent network architecture such as long short-term memory \citep{Hochreiter1997} or \gls{gru} \citep{Cho2014}. The \gls{rnn} receives time windows $\mathbf{x}^n_t$ and transforms them into a single fixed-sized embedding vector $\mathbf{z}^n$ per patient.

Additionally, the simple autoencoder described earlier may be replaced by any number of alternative architectures. For example, we found it beneficial in our experiments to use a \gls{vae} instead \citep{Kingma2014}. In this case, the network learns a probabilistic rather than deterministic patient embedding (commonly parameterised as the means $\mu$ and variances $\sigma^2$ of a $D'$ dimensional multivariate Gaussian distribution with diagonal covariance structure) which in our experiments lead to a smoother, more continuous embedding space. Using a \gls{vae} and changes $L_r$ to

\begin{align}
    L_r = -&  \frac{1}{N}\sum_{n=1}^{N} \log p_{\theta}(\textbf{x}^n | \textbf{z}^n) \\
    & -\frac{1}{2}\sum_{d'=1}^{D'}(1 + log(\sigma_{d'}^2) - \mu_{d'}^2 - \sigma_{d'}^2) \nonumber
    \label{eq:lv}
\end{align}

\noindent
Equation \ref{eq:lrr} may be seen as a special weighted case of $\log p_{\theta}(\textbf{x} | \textbf{z})$ where all input dimensions are modelled as independently Gaussian (with fixed unit variance) or Bernoulli.

\subsubsection{Pre-training of the autoencoder}

As cohorts of interest often consist of much smaller numbers of patients than the total available (e.g., only patients with incident of diabetes), a pre-training step is applied to learn patient embeddings from all available data. This aims to learn a better representation between the wide range of diagnoses, procedures, medications, and laboratory measurements through time before updating the learned patient embeddings on just the cohort of interest.

\subsection{Patient Outcomes}

During pre-training, the autoencoder learns a lower-dimensional representation $\mathbf{z}^n$ in an unsupervised manner. We propose to include a shallow fully-connected layer $h_{\phi}$ that relates $\mathbf{z}^n$ to the risk of experiencing the outcome $y^n$, estimating a scalar risk score $r^n = h_{\phi}(\mathbf{z}^n)$. 

The outcome is then included during training via the additional loss function $L_y$. Depending on the nature of the prediction task, $L_y$ may be chosen as the mean squared error (regression) or binary cross-entropy (classification). For the case of right-censored time-to-event outcomes, we propose to use a loss based on the partial likelihood of the Cox proportional hazards model \citep{Bello2019}, which is defined as

\begin{equation}
    L_y = - \frac{1}{N}\sum^N_{n=1} c^n \left\{ r^n - \log \sum_{j \in R(s^n)} \exp(r^j) \right\}
\label{eq:cox}
\end{equation}

\noindent
where $c$ describes whether an outcome was observed for the patient ($c^n=1$) or if the patient was censored ($c^n=0$) and $R(s^n)$ represents the set of patients still at risk after time $s^n$, i.e., $R(s^n) = \{i~|~i \in \{1, ..., N\}, s^i \geq s^n \}$.

\subsection{Patient Clustering}

Once a patient embedding has been learned (with or without considering the outcome), standard clustering methods may be applied (see for example \cite{Zhang2019}). Alternatively, cluster assignments may be learned simultaneously with the patient embeddings, which allows them to influence the learned embeddings via back propagation and optimise them for clustering. Following \cite{Xie2015-mw}, we introduce a clustering layer that learns the position of $K$ cluster centroids $ \lambda_k \in \mathbb{R}^{D'}$. The probability $q^n_k$ of patient embedding $z^n$ belonging to cluster $k$ can then be calculated via an appropriate kernel, e.g., the density of a Student's t distribution:

\begin{equation}
    q^n_k = \frac{(1+||z^n - \lambda_k||^2)^{-\frac{1}{2}}}{\sum_{k'} (1+||z^n - \lambda_{k'}||^2)^{-\frac{1}{2}}}
\end{equation}

\noindent
Cluster assignments are then iteratively refined \cite{Xie2015-mw}. Since true cluster labels are unknown, we instead use self-training via an auxilliary target distribution $p^n_k$ that emphasise each patient's high confidence clusters 

\begin{equation}
    p^n_k = \frac{(q^n_k)^2 / f_k}{\sum_{k'} (q^n_{k'})^2/f_{k'}}
\end{equation}

\noindent
where $f_k = \sum^N_{n=1} q^n_k$ is used to normalise cluster frequencies \citep{Xie2015-mw}. By penalising large differences between $q^n_k$ and $p^n_k$, the network is incentivized to pull patient embeddings towards a single (closest) centroid. The corresponding clustering loss $L_c$ is  defined as

\begin{equation}
    L_{c} = \kldiv{P}{Q} = \frac{1}{N}\sum^N_{n=1} \sum^K_{k=1} p^n_k \log \frac{p^n_k}{q^n_k}
\label{eq:lc}
\end{equation}

\noindent
where $\kldiv{P}{Q}$ indicates the \gls{kl} divergence between distributions $P$ and $Q$. See \cite{Xie2015-mw} for a more detailed discussion.

\subsection{Evaluation Metrics}

\subsubsection{Cluster Similarity}

The \gls{ari} is used to measure the similarity between two sets of data clusters. The Rand index is defined as,

\begin{equation}
    RI = \frac{a + b}{{N \choose 2}},
\label{eq:RI}
\end{equation}

\noindent
where, for two sets of clusters $C$ and $K$, $a$ represents the number of pairs of elements in the same cluster in $C$ and $K$, and $b$ represents the number of pairs of elements in different clusters in $C$ and $K$. The \gls{ari} ensures random label assignment have a score close to zero and is defined as,

\begin{equation}
    ARI = \frac{RI - \mathbb{E}(RI)}{max(RI)-\mathbb{E}(RI)},
\label{eq:ARI}
\end{equation}

\noindent
where $\mathbb{E}(RI)$ is the expected RI of random assignments.


\subsubsection{KM Curves and Log Rank Test}

\gls{km} curves are used to evaluate the time-to-event within each of the discovered clusters \citep{Kaplan1958}. It measures the fraction of patients $\hat{P}(s)$ who have not experienced the event of interest by time $s$, defined as

\begin{equation}
\hat{P}(s) = \prod_{i: s_i \leq s} \left( 1 - \frac{d_i}{n_i}\right)
\label{eq:km}
\end{equation}

\noindent
where $d_i$ is the number of events that happened at time $s_i$, and $n_i$ is the number of patients who were still observed at that time. In the absence of competing risks, the crude incidence curve of the outcome can be calculated as $1-KM$ \citep{Austin2016}. \gls{km} and crude incidence curves allow for an intuitive graphical comparison of the average risk in each cluster. We used log rank tests to formally compare clusters for differences in outcome risk \citep{Harrington1982}. In short, the log rank test assesses the null hypothesis $H_0$ of equal \gls{km} curves across all discovered clusters. Larger values of the test statistic therefore indicate more separated curves. Note, however, that the test statistic may be driven by a large difference of only a single cluster and therefore needn't indicating separation between all clusters.

\section{Data}

We evaluated our model on two datasets: a synthetic \gls{ehr} dataset with known clusters and a real world \gls{ehr} dataset of diabetes patients from which the model is used to derive clinical insights.

\subsection{Synthetic Data}

We demonstrate the idealised behaviour of our proposed model within synthetic data with a known data structure. We simulated three types of clusters: unsupervised clusters, outcome clusters,  and combined clusters. 
Unsupervised clusters, share similarities in the input space but were not associated with the outcome.
These clusters are susceptible to data bias (e.g., similarities in patient trajectories due to local hospital guidance) and therefore might be of less scientific interest.
Outcome clusters share the same risk of developing the event of interest but have no associated feature combinations.
Combined clusters, on the other hand, represent groups of patients which share feature combinations in the input space that are associated with a higher or lower risk of developing the event of interest (e.g., a combination of factors that increase the risk of death). 
We hypothesise that these clusters are more clinically relevant and and their identification is the goal of this study. 

We chose the variance of unsupervised clusters such that it was larger than that of combined clusters. 
This ensured that they were favoured by purely unsupervised clustering methods (e.g., \gls{pca} k-means or \gls{dec}), whereas semi and supervised methods (e.g., \gls{rsf}, AC-TCP) are expected to find the simulated outcome clusters. 
However, as discussed earlier, the latter disregard different patient trajectories in the input space that lead to similar outcomes. In order to show that --- depending on the weighting of the loss functions --- our proposed model can also recover the specific trajectories that lead to outcomes, we further split the outcome clusters into subgroups that shared the same outcome distribution but a different covariate distribution. 

The synthetic dataset is generated for $P=60,000$ patients with the details of synthetic data generation shown in Appendix \ref{apd:synthetic}. 

\subsection{Real World Data}
Data was collected by the Oxford University Hospitals NHS Foundation Trust between August 2014 and March 2020 as part of routine care. The longitudinal secondary care \gls{ehr} includes demographic information (i.e. sex, age), admission information (start/end dates, discharge method/destination, admission types - e.g. in-patient, outpatient, emergency department), ICD-10 coded diagnoses, OPCS-4 coded procedures, medications as British National Formulary (BNF) codes (prescribed both during visits and take-home), and laboratory measurements (e.g. blood and urine tests). Diagnosis codes could either appear in the data as a primary (indicating the primary reason for the hospital admission) or secondary diagnosis (further present comorbidities). While the majority of these are binary or categorical features, laboratory values are continuous. 

Data from 493,470 patients was available for pre-train the \gls{rnn} autoencoder for the initial patient trajectory embedding.
Sequential data is created for all patients by grouping features in to time windows or bins.
Note, even though time was not explicitly treated as a covariate, windows with no data were not removed from the sequence such that model can estimate the time difference between irregular sampled observations.
Each trajectory of a patient $n$ was divided into non overlapping time windows $x^n_{t}$ of 90 days, with $t$ being the time index. 
As the data spanned more than five years, this resulted in up to $t_{max}= 22$ windows per patient. 
Whereby features with a occurrence of $<1\%$ were removed.

Features were extracted per time window if data was present. The binary features (primary and secondary diagnosis, procedures and medication codes) were included using multi-hot encoding. Laboratory values within a time window $x_t$ were encoded using 6 features: \textit{min}, \textit{max}, \textit{mean}, \textit{median absolute deviation (MAD)} as well as the \textit{last} value within the time window and \textit{number of occurrences} per time window. The laboratory values were normalised using rank normalization \citep{Qiu2013}, where values for a given laboratory measurement were ranked according to all values in the cohort and then the ranks were normalized to the range $[0,\,1]$.
Missing binary features within a window are filled with zeros, missing continuous features are filled with $-0.1$, a value outside of the possible range of the normalised values. 
Time windows with no data were filled with an \textit{empty} vector consisting in zeros for the binary features, and $-0.1$ for the continuous features.
To reduce the impact of missing data points or empty time windows, these values were masked in the reconstruction loss while training the \gls{vae}.

After filtering, the total number of different features can be broken down into this feature type categories: 286 primary diagnosis codes, 351 secondary diagnosis codes, 175 procedure codes, 122 medication types and 55 laboratory values. A summary of the full cohort feature types and average lengths of trajectories are shown in Table \ref{tab:cohort_stats}.

\begin{table}[!h]

  \resizebox{\columnwidth}{!}{
  \begin{tabular}{l|c|c}
  \toprule
    & Diabetes Cohort &  Full Cohort \\
  \midrule
    \# of patients           & 29,229 & 493,470 \\
  \midrule
    Data Windows \#          & 272,390 & 2,543,106\\
    Data Windows Avg. per Patient & 9.3 & 5.2 \\
  \midrule
    Primary ICD-10 \#  Unique          & 286 & 286 \\
    Primary ICD-10 Avg. per Window & 0.29 & 0.17 \\
    Primary ICD-10 Frac. of Windows & 0.23 & 0.14 \\
  \midrule
    Secondary ICD-10 \#  Unique          & 351 & 351 \\
    Secondary ICD-10 Avg. per Window & 1.9 & 0.8 \\
    Secondary ICD-10 Frac. of Windows & 0.28 & 0.19 \\
  \midrule
    OPCS-4 \#  Unique        & 175 & 175 \\
    OPCS-4 Avg. per Window  & 0.82 & 0.57 \\
    OPCS-4 Frac. of Windows  & 0.30 & 0.24 \\
  \midrule
    Medications \#  Unique           & 122 & 122 \\
    Medications Avg. per Window & 3.0 & 1.6 \\ 
    Medications  Frac. of Windows  & 0.24 & 0.21 \\
  \midrule
    Lab Measurements \#  Unique           & 55 & 55 \\
    Lab Measurements Avg. per Window & 16.1 & 13.9 \\
    Lab Measurements  Frac. of Windows  & 0.89 & 0.85 \\
  \bottomrule
  \end{tabular}
  }
  \caption{\label{tab:cohort_stats} Statistical description of the cohorts and trajectories used. \# refers to the number of patients or unique features of the different data types present. The Avg. per Window, refers to the average number of features from a given type present in a window with data. Frac. of Windows refers to the fraction of windows with data that contains at least one of the corresponding feature type.}
\end{table}

\subsubsection{Diabetes Cohort}

A cohort of $29,299$ diabetes patients were selected from the full cohort to test the model on a specific cohort. Patients were included in the cohort if they had at least one primary or secondary diagnoses of diabetes, their first diagnosis of diabetes is used as an index date. Unlike the full cohort, where the trajectories are unaligned. The diabetes patient trajectories are aligned at the window containing the index event, ensuring all patients have the same number of windows (including empty windows) and the index event occurs in the same window in each patient. 
We investigate as a clinical outcome the risk of future cardiovascular events, of which diabetes is a risk factor. 
Time-to-event labels were defined as the time from index date of first diabetes diagnosis to the first occurrence of stroke, myocardial infarction, or other bleeding event. A summary of the average number of feature types and windows with data for the diabetes cohort is shown in Table \ref{tab:cohort_stats}, with a detailed feature summary in the Appendix \ref{apd:datasummary}.

\section{Results}

\subsection{Experiment Setup}

\subsubsection{Baseline Methods}

\gls{pca} with k-means clustering is used as an unsupervised clustering baseline. As k-means clustering is a distance based clustering method which does not perform well in high dimensional spaces, the 256 dimensional patient embedding is first reduced using \gls{pca} by taking the first five principal components and applying k-means clustering to these components.

\glspl{rsf} are used as a supervised clustering baseline, to find clusters of patients who share similar time to events. A single tree of depth four is trained on 75\% of features, resulting in 16 possible risk scores (one at each leaf node). This is repeated ten times with random subsets of features, resulting in each patient having ten risk scores. K-means clustering is then applied to the risk scores to obtain the final supervised clusters.

\subsubsection{Model Training}

The initial \gls{rnn} autoencoder model, trained on the full cohort of $493,479$ patients, was trained for $350$ epochs with a batch size of $4,096$ and a learning rate of $2\times10^{-3}$ using gradient descent with an Adam optimiser. A weight decay of $1\times10^{-6}$ is used for regularisation, and dropout used between the \gls{gru} layers ($p=0.1$). The output dimension of the fully connected encoder layers was $256$, with the hidden state of the \gls{gru} having dimensions of $256$. The model architecture is described in more detail in Appendix \ref{apd:architecture}. 

The proposed \gls{lps-co} model, trained on the diabetes cohort of $29,299$ patients, was trained for $25$ epochs with a batch size of $256$ and a learning rate of  $1\times10^{-3}$ using gradient descent with an Adam optimiser. A weight decay of $1\times10^{-6}$ is used for regularisation, and dropout used between the \gls{gru} layers ($p=0.1$). Model dimensions remain the same as the initial \gls{rnn} autoencoder training. Hyperparamters are selected to ensure losses are converging, although no formal optimisation was applied. All models were built using PyTorch.

Three versions of the proposed \gls{lps-co} model are used with different loss weights (Equation \ref{eq:loss}) for reconstruction loss, $w_r$, and outcome loss, $w_y$: no outcome loss ($w_r=0.5$, $w_y=0$), no reconstruction loss ($w_r=0$, $w_y=1$), and both reconstruction and outcome loss ($w_r=0.05$, $w_y=1$), these weights are chosen to ensure the losses are of similar magnitudes when combined, they have not been optimised and are left to the user depending on model requirements. In all models the KL divergence loss weight, $w_{kl}$, is set to $1\times10^{-5}$ and the clustering loss weight is set to $0.25$.

\subsection{Synthetic Data Results}

To validate the proposed model and evaluate the combination of reconstruction and outcome loss, the model was applied to the synthetic dataset. The three versions of the \gls{lps-co} model with different loss weights were trained on the synthetic data. In addition to the proposed model, PCA k-means was trained as a baseline unsupervised clustering model, a random survival forest was trained as a baseline supervised clustering model, and an AC-TCP model proposed by \cite{Lee2020} was trained as a state-of-the-art comparison. All models were trained two times, once to find three clusters and once to find six clusters.

Table \ref{tab:exp1a} shows the \gls{ari} scores comparing the discovered clusters of each models to the true labels of the unsupervised, outcome, and combined clusters of the synthetic data. \gls{pca} k-means and \gls{lps-co} with no outcome loss were able to perfectly find the unsupervised clusters when $k=3$ ($ARI=1$), and could not find the outcomes clusters for $k=3$ or combined clusters for $k=6$ well. The random survival forest, AC-TCP, and \gls{lps-co} with no reconstruction loss accurately found the outcomes clusters for $k=3$. The discovered clusters for $k=6$ shared some similarities with the true combined clusters with \gls{ari} scores of $0.50$, $0.50$, and $0.49$ respectively.

The \gls{lps-co} using combined reconstruction loss and outcome loss obtained the highest \gls{ari} score for the combined clusters for $k=6$. The \gls{ari} score of $0.78$ was higher than that of both the reconstruction clusters ($0.02$) and the outcomes clusters ($0.69$) indicating the model is able to ignore the large data biases in the data whilst focusing on the less prominent patterns in the input space associated with different outcomes.

\begin{table}[!h]
\resizebox{\columnwidth}{!}{
\begin{tabular}{l|ccc|ccc}
\hline
 & & k=3 & & & k=6 & \\
\hline
                  & Unsup. & Outcome & Combined & Unsup. & Outcome & Combined \\
\hline
PCA k-means       & 1.00 & 0.00 & 0.00 & 0.69 & 0.08 & 0.13 \\
RSF               & 0.00 & 0.96 & 0.55 & 0.02 & 0.71 & 0.50 \\
AC-TCP            & 0.00 & 1.00 & 0.57 & 0.02 & 0.84 & 0.50 \\
\gls{lps-co}  & & & \\
($w_r=0.5$, $w_y=0$)    & 1.00 & 0.00 & 0.00 & 0.76 & 0.04 & 0.10 \\
\gls{lps-co} & & & \\ 
($w_r=0$, $w_y=1$)   & 0.00 & 1.00 & 0.57 & 0.00 & 0.77 & 0.49 \\
\gls{lps-co}  & & & \\ 
($w_r=0.05$, $w_y=1$) & 0.00 & 1.00 & 0.57 & 0.02 & 0.69 & \textbf{0.78} \\
\hline
\end{tabular}
}
\caption{\label{tab:exp1a}Adjusted Rand index scores for baseline and proposed \gls{lps-co} models on synthetic laboratory measure data with known unsupervised, outcome, and combined clusters.}
\end{table}

\subsection{Diabetes Dataset Results}

The model is also validated on real world data with the cohort of diabetes patients and using the time to first cardiac event as the outcome. An initial patient embedding is trained on the full dataset of $493,470$ patients using only the reconstruction loss before further training of the proposed clustering model on the diabetes cohort. Three versions of the model were trained on the diabetes cohort: without outcome loss, without reconstruction loss, and with combined reconstruction and outcome loss. The models were trained multiple times to find clusters from $k=2$ to $k=7$ resulting in $18$ different scenarios. Models were trained five times on $80\%$ of the data within each scenario and the results averaged to determine the robustness of the models.

\begin{table}[!h]
\resizebox{\columnwidth}{!}{
\begin{tabular}{lccc}
\hline
Clusters & Recon.-Combined & Outcome-Combined & Recon.-Outcome  \\
\hline
2 & $0.07\pm0.08$ & $0.18\pm0.11$ & $0.06\pm0.09$ \\
3 & $0.31\pm0.22$ & $0.30\pm0.23$ & $0.13\pm0.16$ \\
4 & $0.10\pm0.02$ & $0.21\pm0.11$ & $0.07\pm0.05$ \\
5 & $0.15\pm0.02$ & $0.22\pm0.13$ & $0.07\pm0.03$ \\
6 & $0.14\pm0.07$ & $0.24\pm0.16$ & $0.09\pm0.05$ \\
7 & $0.14\pm0.05$ & $0.27\pm0.07$ & $0.07\pm0.02$ \\
\hline
\end{tabular}
}
\caption{\label{tab:ARI}Adjusted Rand index scores between pairs of \gls{lps-co} clusters from different loss weights, showing similarities between the discovered clusters for each k.}
\end{table}

Training without outcome loss, the models have no information about the time to cardiac event outcome, thus the clusters can only be driven by the patient trajectories up to the event of first diabetes diagnosis. Similarly, training without reconstruction loss, the models try and cluster patients who have differing outcomes and not similar trajectories. As we want to find patients who share similar trajectories and have different outcomes, ideally the model with combined losses shares information with both the clusters driven by the trajectories and the clusters driven by outcomes.

Table \ref{tab:ARI} shows the mean and standard deviation of \gls{ari} scores comparing clusters found using reconstruction loss with combined loss, outcome loss with combined loss, and reconstruction loss with outcome loss. The \gls{ari} scores between the reconstruction and outcome loss clusters are low (a maximum of $0.13\pm0.16$ for $k=3$), indicating little similarity between the discovered clusters. This is as expected due to the differing focuses on trajectories and outcomes. The \gls{ari} scores between the combined clusters and both the reconstruction and outcome clusters are higher in all cases, showing the combined loss model is learning from both trajectories and outcomes. The \gls{ari} scores between combined loss clusters and outcome loss clusters are generally higher than the scores between combined loss clusters and reconstruction loss clusters, suggesting the combined losses focus more on the outcomes. The strength of the focus can be altered by changing the weights of each of the losses. Additional metrics are shown in the Appendix \ref{apd:clustermetrics}.

\begin{table}[!h]
\resizebox{\columnwidth}{!}{
\begin{tabular}{lccc}
\hline
 & \gls{lps-co} & \gls{lps-co} & \gls{lps-co} \\
Clusters & ($w_r=0.5$, $w_y=0$) & ($w_r=0$, $w_y=1$) & ($w_r=0.05$, $w_y=1$) \\
\hline
2 & $345\pm171$ & $950\pm793$ & $825\pm514$ \\
3 & $766\pm112$ & $7483\pm2691$ & $5366\pm2520$ \\
4 & $1169\pm217$ & $7779\pm4773$ & $4029\pm4732$ \\
5 & $1900\pm343$ & $6637\pm3846$ & $2657\pm884$ \\
6 & $1190\pm151$ & $5958\pm2129$ & $7527\pm3036$ \\
7 & $1961\pm447$ & $10770\pm2367$ & $8575\pm1955$ \\
\hline
\end{tabular}
}
\caption{\label{tab:logrank}Log rank test statistic between reconstruction ($w_r=0.5$, $w_y=0$), outcome ($w_r=0$, $w_y=1$), and combined ($w_r=0.05$, $w_y=1$) loss clusters, showing separation of outcomes between the discovered clusters for each k.}
\end{table}

\begin{figure*}[ht]
\centering
\includegraphics[width=140mm]{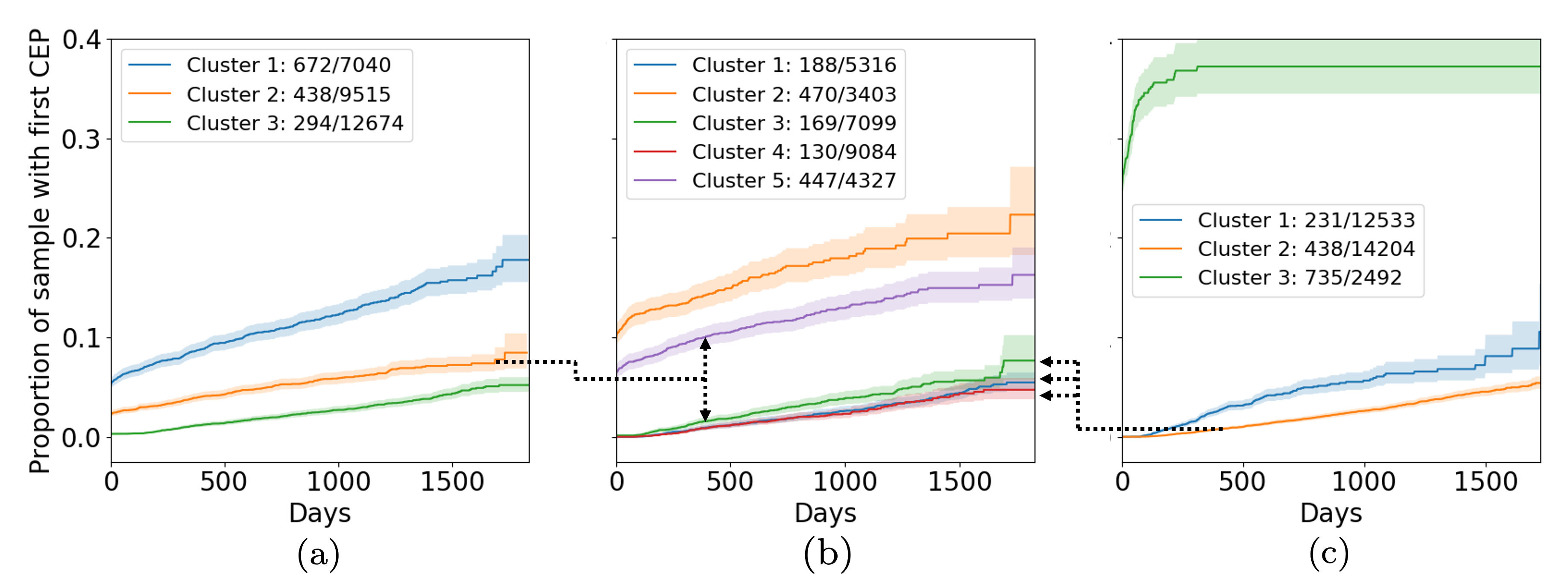}
\caption{\label{fig:KMcurves}Kaplan-Meier curves for (a) Reconstruction loss clusters ($k=3$, $w_r=0.5$, $w_y=0$), (b) combined loss clusters ($k=5$, $w_r=0.05$, $w_y=1$), and (c) Outcome loss clusters ($k=3$, $w_r=0$, $w_y=1$). For the values (X/Y), X shows the number of patients in the cluster who have a cardiac outcome, and Y shows the total patients in the cluster. Arrows indicate examples of how patients in clusters from the $k=3$ reconstruction loss only and outcome loss only move to new clusters in the $k=5$ combined loss model.}
\end{figure*}

Figure \ref{fig:KMcurves} (a) and (c) show the \gls{km} curves for the clusters found using the reconstruction loss only, combined loss, and outcome loss only models for $k=3$. The curves estimate the time to first cardiac event for the patients in each cluster. We see that for reconstruction loss only in Figure \ref{fig:KMcurves} (a) the curves are less separable, with the curves for the outcome loss only in Figure \ref{fig:KMcurves} (c) most separable. Using combined losses in Figure \ref{fig:KMcurves} the separation of the \gls{km} curves are between the two other models. This is quantified in Table \ref{tab:logrank} where the mean and standard deviation of the multivariate log rank test statistic can be seen for $k=2$ to $k=7$ for each model. In all cases, the test statistic is highest for the outcome loss only model, indicating the highly differing \gls{km} curves, and lowest for the reconstruction loss only, indicating similar \gls{km} curves in each cluster. Again, the combined loss model is intermediate, showing it is combining both outcome and trajectory information. 

\begin{table}[!h]
\resizebox{\columnwidth}{!}{
\begin{tabular}{lccccc}
\hline
Clusters & Comb. 1 & Comb. 2 & Comb. 3 & Comb. 4 & Comb. 5 \\
\hline
Recon. 1 & 0.04 & 0.29 & 0.40 & 0.15 & 0.12 \\
Recon. 2 & 0.01 & 0.13 & 0.42 & 0.10 & 0.34 \\
Recon. 3 & 0.39 & 0.01 & 0.02 & 0.56 & 0.02 \\
\hline
Outcome 1 & 0.01 & 0.15 & 0.32 & 0.40 & 0.13 \\
Outcome 2 & 0.37 & 0.03 & 0.21 & 0.27 & 0.12 \\
Outcome 3 & 0.00 & 0.43 & 0.07 & 0.10 & 0.40 \\
\hline
\end{tabular}
}
\caption{\label{tab:cluster_comp}The fraction of patients from each cluster of the combined loss model (k=5) in each cluster of the reconstruction loss only and outcome loss only models (k=3).}
\end{table}

We also investigate how patients in clusters found from trajectories (reconstruction loss only) and outcomes (outcome loss only) for small numbers of clusters ($k=3$) split and are distributed through higher number of clusters ($k=5$) using a combined loss model. Table \ref{tab:cluster_comp} shows the distributions of patients from each of the clusters for $k=3$ within the clusters for the combined loss model using $k=5$. For example, the patients in cluster $3$ of the outcome loss only model ($k=3$) are now primarily distributed between clusters $2$ ($43\%$) and $5$ ($40\%$) of the combined loss model ($k=5$).

Two cases on how clusters of patients are split going from $k=3$ to $k=5$ can be highlighted. Firstly, patients in cluster $2$ using reconstruction loss only $k=3$, are primarily assigned to clusters $3$ ($42\%$) and $5$ ($54\%$) from the combined loss model ($k=5$). From Figure \ref{fig:KMcurves} (b) we can see clusters $3$ and $5$ have different outcomes, yet we know the patients share similar trajectories as they are assigned to the same clusters in the reconstruction only model ($k=3$). This indicates the combined loss model is able to find clusters of patients with similar trajectory, but different outcomes.

For the second case, we see that patients in cluster $2$ of the outcome loss only model ($k=3$) are mainly distributed between clusters $1$ ($37\%$), $3$ ($21\%$), and $4$ ($27\%$) of the combined loss model ($k=5$). From Figure \ref{fig:KMcurves} (b), we see clusters $1$, $3$, and $4$ share similar outcomes (as in cluster $2$ in Figure \ref{fig:KMcurves} (c)), therefore must have differing trajectories to be separate clusters. This further indicates that the combined loss model separates patients both on trajectory and outcome.

\section{Discussion}

We have developed a novel \gls{rnn} autoencoder model to cluster patient trajectories from \gls{ehr} data using a combination of losses. We combine the more standard reconstruction loss with a time-to-event loss to discover clusters of patients with both different trajectories and outcomes and evaluated it on a synthetic and real world dataset. 

Our evaluation using synthetic data showed that our approach was able to find clusters based on the trajectories and outcomes (Table \ref{tab:exp1a}) by adjusting the weight parameters $w_r$ and $w_y$ of the loss functions. 
However, one limitation is that it is unclear how these weights should be determined when working with real \gls{ehr} data with unknown underlying clusters. 
As there are various solutions possible, we did not perform a detailed clinical enrichment analysis of the identified clusters in the diabetes dataset.
Future work needs to investigate how these parameters can be optimised for a specific application. 
For instance, our approach can be used to identify patient cohorts which are more suited for a clinical trial (e.g. having a higher likelihood of a clinical outcome which can reduce the trial duration).
In such a scenario, further criteria such as number of  features required to define a cluster (number of inclusion \& exclusion criteria) and clinical interpretability of these could be included.

Another challenge is that time is not directly considered in the model. 
The temporal resolution of our approach was $90$ days (size of a single time window). The optimal temporal granularity depends on the specific clinical question and will influence the cluster outcome. 
Approaches such as \cite{Baytas2017} which integrate time directly should be investigated further.





\acks{
This work uses data provided by patients collected by the NHS as part of their care and support. We believe using the patient data is vital to improve health and care for everyone and would, thus, like to thank all those involved for their contribution. The data were extracted, anonymised, and supplied by the Trust in accordance with internal information governance review, NHS Trust information governance approval, and the General Data Protection Regulation (GDPR) procedures outlined under the Strategic Research Agreement (SRA) and relative Data Processing Agreements (DPAs) signed by the Trust and Sensyne Health plc.
 
This research has been conducted using the Oxford University Hospitals NHS Foundation Trust Clinical Data Warehouse, which is supported by the NIHR Oxford Biomedical Research Centre and Oxford University Hospitals NHS Foundation Trust. Special thanks to Kerrie Woods, Kinga Várnai, Oliver Freeman, Hizni Salih, Steve Harris and Professor Jim Davies.
}

\bibliography{references}

\appendix

\section{Generating Synthetic Data}
\label{apd:synthetic}

\begin{figure*}[!h]
\centering
\subfigure[First two principal components of noise features coloured by data bias cluster label]{
  \includegraphics[width=50mm]{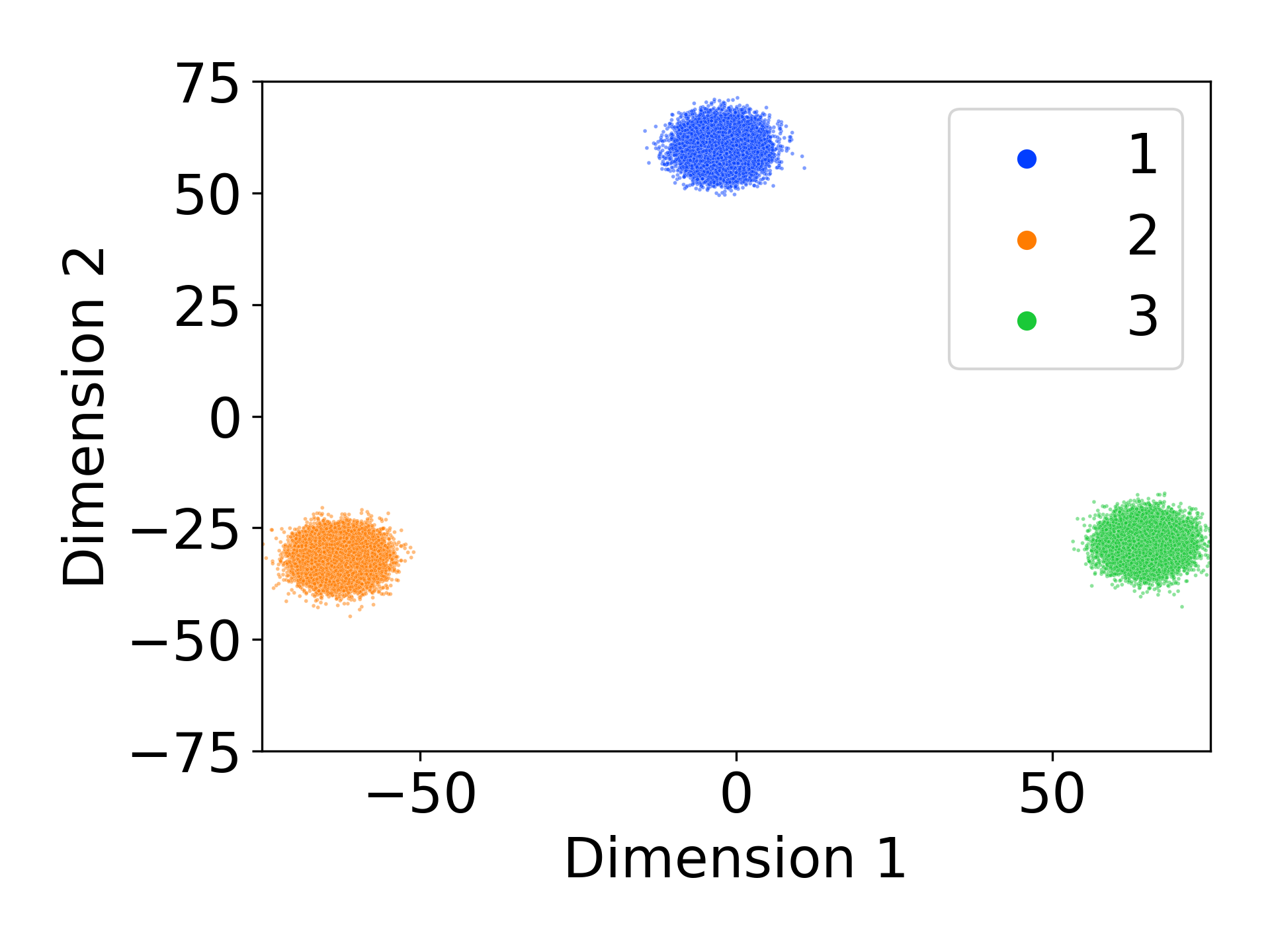}
  \label{fig:syntheticA}
}
\subfigure[First two principal components of clinical features coloured by outcome cluster label]{
  \includegraphics[width=50mm]{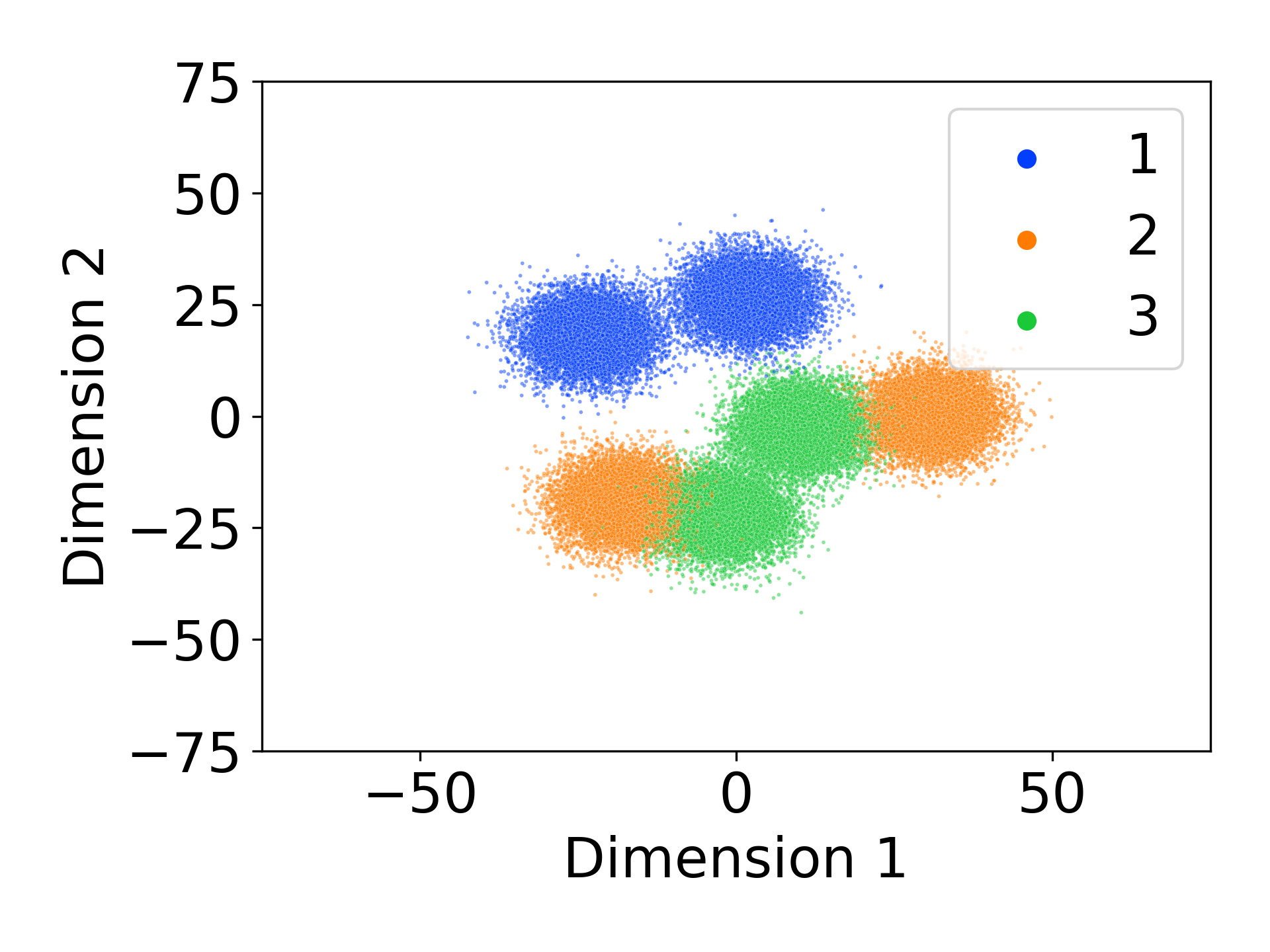}
  \label{fig:syntheticB}
}
\subfigure[First two principal components of clinical features coloured by combined cluster label]{
  \includegraphics[width=50mm]{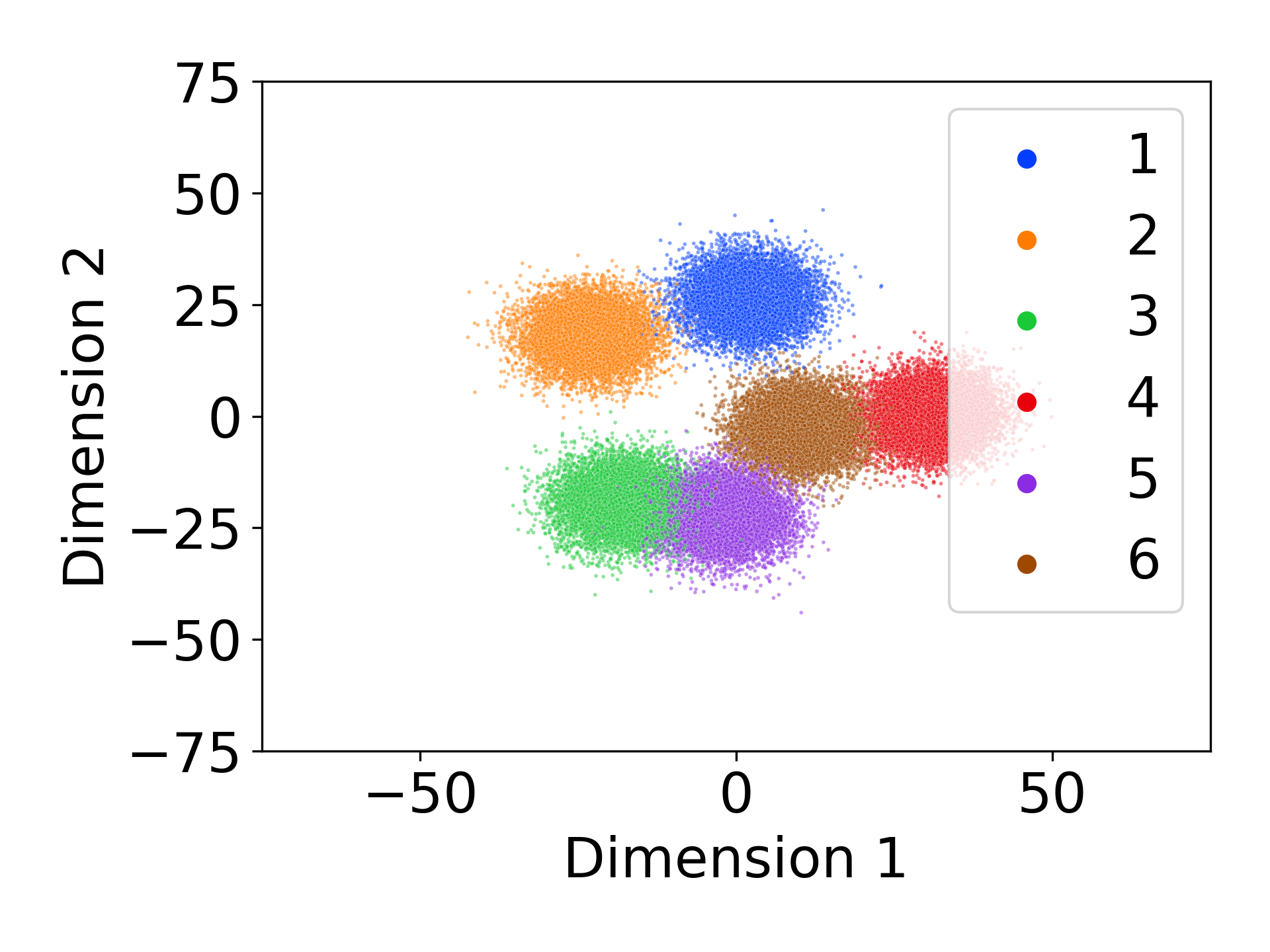}
  \label{fig:syntheticC}
}
\hspace{0mm}
\subfigure[Kaplan-Meier curves of the noise clusters]{
  \includegraphics[width=50mm]{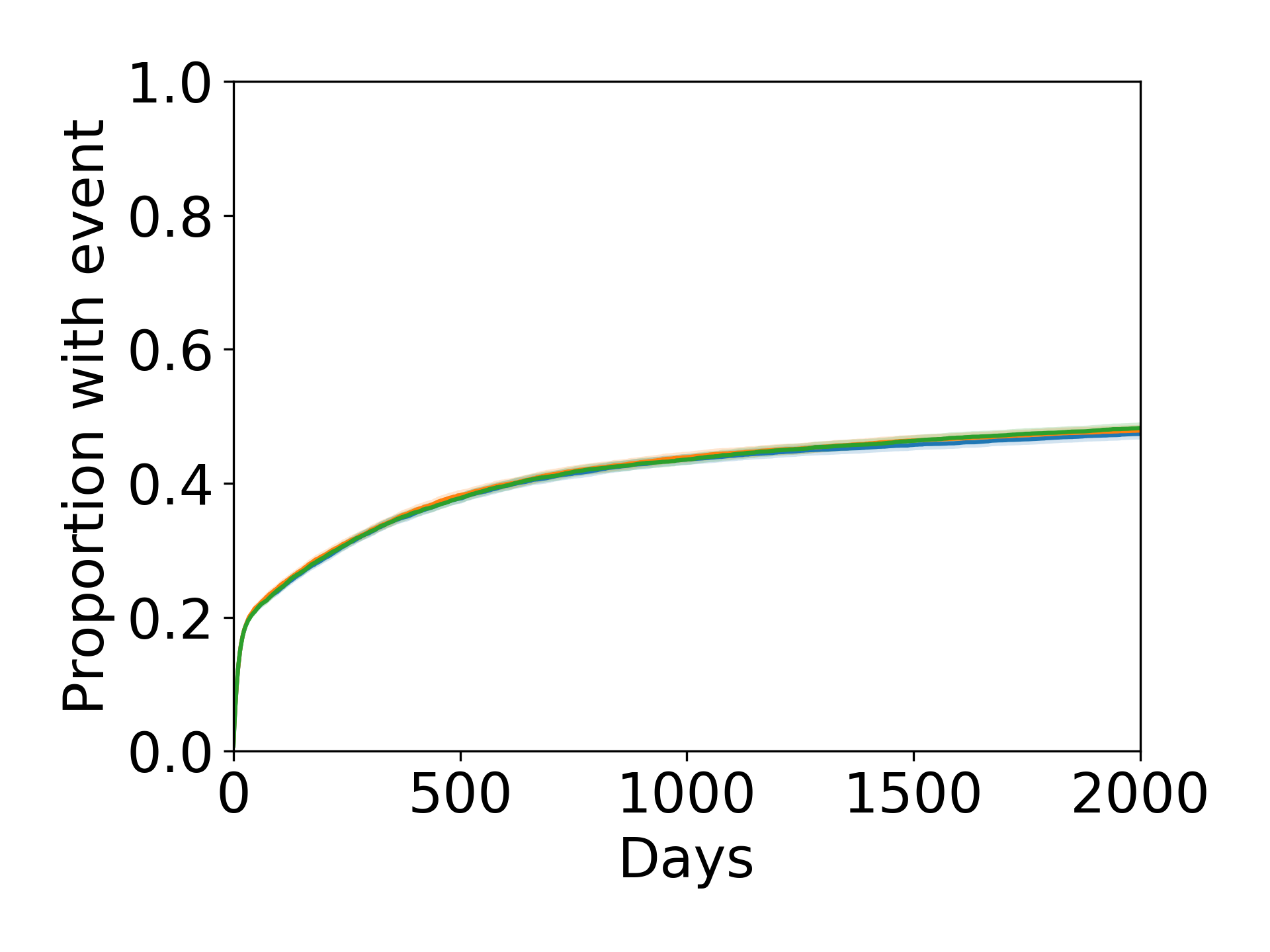}
  \label{fig:syntheticD}
}
\subfigure[Kaplan-Meier curves of the outcome clusters]{
  \includegraphics[width=50mm]{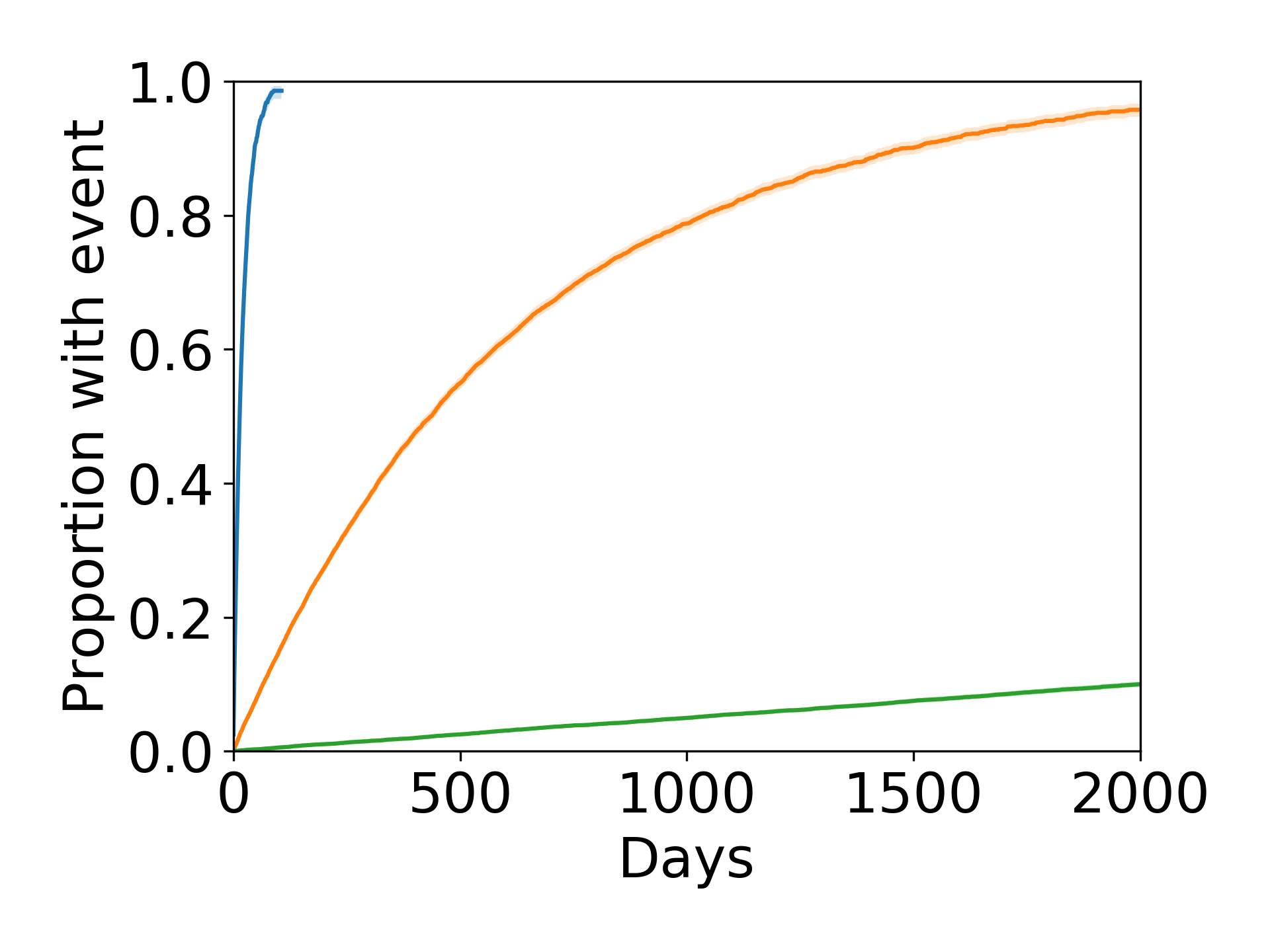}
  \label{fig:syntheticE}
}
\subfigure[Kaplan-Meier curves of the clinical clusters]{
  \includegraphics[width=50mm]{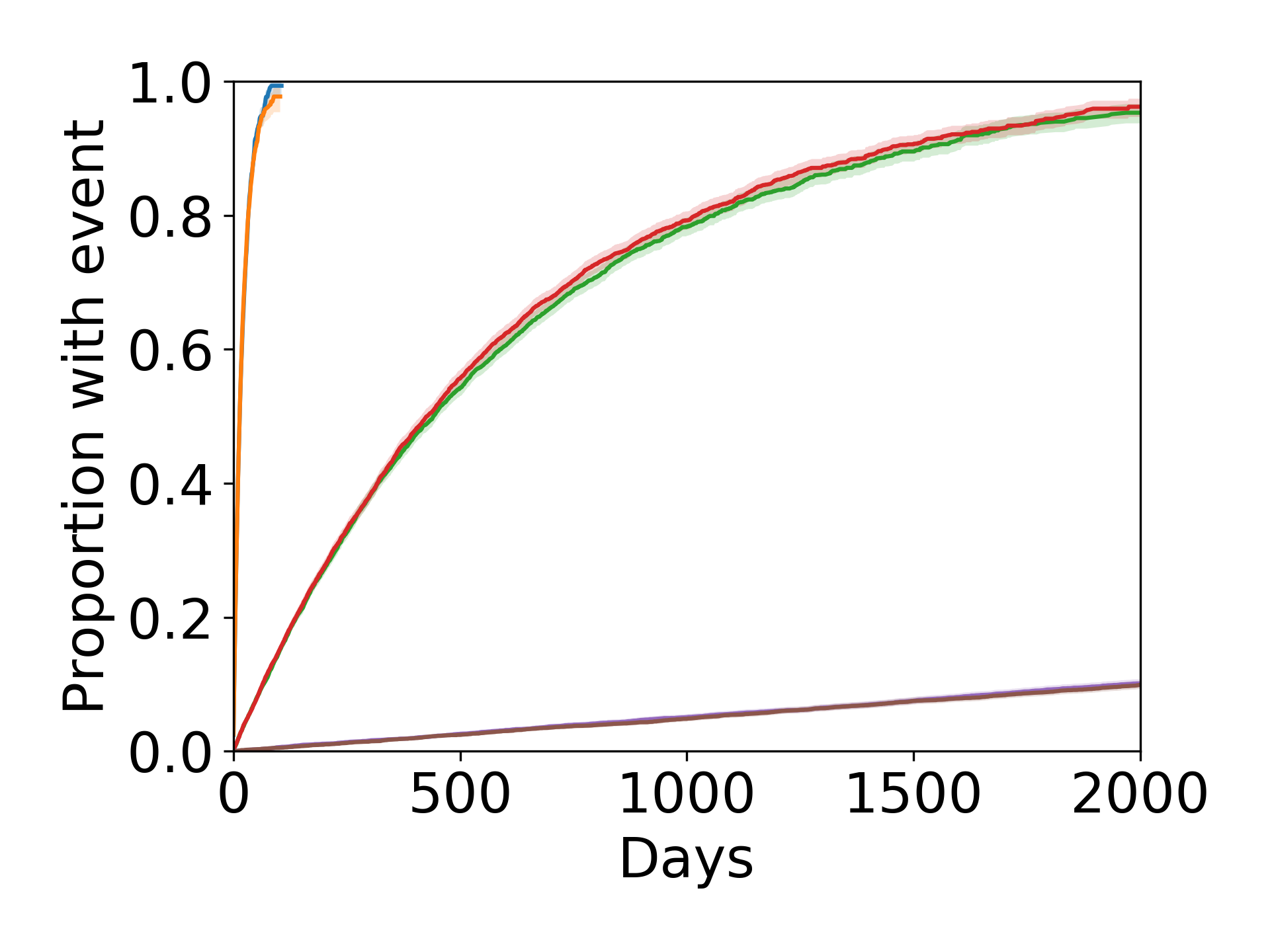}
  \label{fig:syntheticF}
}
\caption{Principal component plots of the data bias and combined features and the Kaplan-Meier curves of the three known cluster groups.}
\label{fig:synthetic}
\end{figure*}

The synthetic data is generated with the following steps:
\begin{itemize}
    \item The number of noise clusters, $K_{noise}=3$, is specified along with the number of synthetic features which contribute to these clusters, $N_{noise}=200$. Features are sampled from isotropic Gaussian distributions with standard deviation $C_{std}=3$ and with cluster centroids generated at random within a bounding box, $(centre_{min}=-10, centre_{max}=-10)$. The generated features are continuous and represent synthetic laboratory measures. In order to generate synthetic binary features (eg. diagnosis codes), synthetic continuous features can be passed through a min max scaler and rounded to zero or one.
    
    \item The number of outcome clusters, $K_{outcome}=3$, is specified along with minimum and maximum time to events ($TTE_{min}=10$, $TTE_{max}=10,000$). The time to events are generated by sampling from exponential distributions with the scale of the distribution for each cluster one of the values log-spaced between $TTE_{min}$ and $TTE_{max}$ with $K_{outcome}$ steps. Censoring of events is sampled randomly from a uniform distribution ($p=0.5$), with all time to events over a maximum threshold ($2,000$) set to this max value and censored.   
    
    \item The number of combined clusters, $K_{combined}=6$, is set to twice the value of $K_{outcome}$ and the number of synthetic features which correlate with outcomes, $N_{outcome}=200$, is specified. Each outcome cluster is randomly split in half to create the combined cluster labels. Features corresponding to the combined clusters are generated using the same method as the feature bias cluster, with smaller distances between the cluster centroids ($C_{std}=5$, $(centre_{min}=-5, centre_{max}=-5)$.
    
\end{itemize}

\noindent
Figure \ref{fig:synthetic} shows the different synthetic clusters and outcomes (as Kaplan-Meier curves). Figure \ref{fig:syntheticA} shows the first two principal components of PCA applied to the noise features, with the colours representing the known cluster labels. Figures \ref{fig:syntheticB} and \ref{fig:syntheticC} show the first two principal components of PCA applied to the clinical features, with the colours representing the known outcome cluster labels and clinical cluster labels respectively. Figures \ref{fig:syntheticD}, \ref{fig:syntheticE}, and \ref{fig:syntheticF} show the Kaplan-Meier curves for the outcomes of the known noise clusters, outcome clusters, and clinical clusters respectively.

\newpage

\section{Experiment architecture}
\label{apd:architecture}

\begin{figure*}
\centering
\includegraphics[width=140mm]{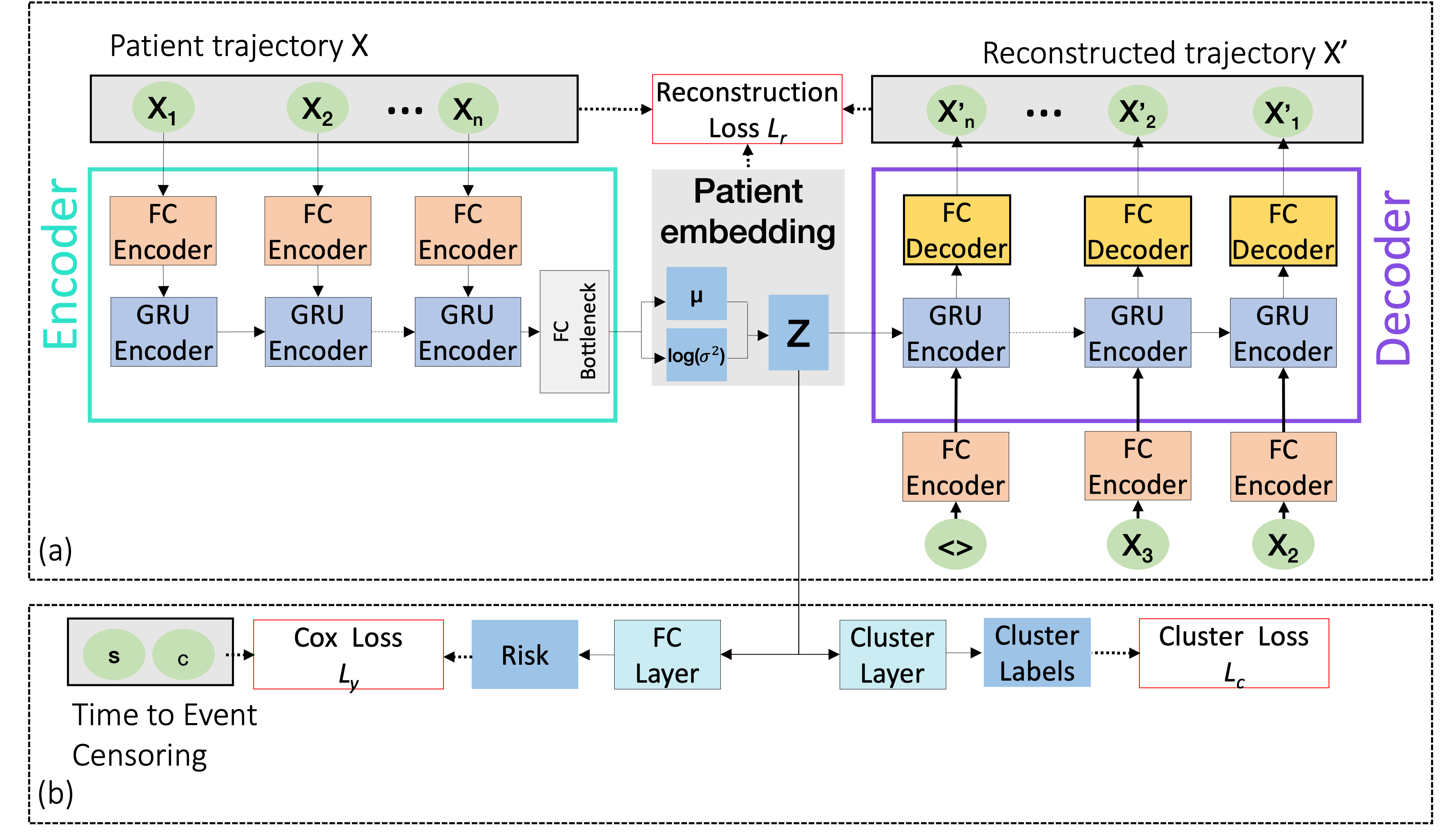}
\caption{Schematic diagram of the RNN autoencoder. (a) shows the standard RNN autoencoder to obtain patient embeddings from trajectories. (b) shows the addition layers and loss functions to update the embedding and obtain cluster assignments.}
\label{fig:RNNAuto}
\end{figure*}

Figure \ref{fig:RNNAuto} shows a schematic diagram of the RNN autoencoder used to create the patient trajectory embedding in (a), using a reconstruction loss and KL divergence loss for a variational RNN autoencoder. Section (b) shows the addition clustering loss and outcome loss incorporated into the \gls{lps-co} model to find clusters of patients who have differences in both trajectories and outcomes.

Our proposed model, a variational RNN autoencoder, is illustrated in Figure \ref{fig:RNNAuto} (a). 
The encoder consists of a feature embedding layer which transforms the features of time window $\textbf{x}_i$ into a 256 fixed-sized embedded vector using two fully connected layer with ReLU activation function.
Each time window is then passed into a two layer bidirectional GRU using a dropout of 0.1 between the layers. The last hidden layer output has a dimension of 256x4 tensor (two directions times two layers). This tensor is aggregated into a 256 vector using a fully connected layer with a ReLU activation function. The aggregated vector is then feed to two separated fully connect layers producing two 256 vectors, representing the means and the log variances of the normal distributions from which the patient embedding, $Z$, is sampled using the re-parameterization trick \cite{Kingma2014}.

The decoder aims to predict the trajectory sequence in reverse order and uses teacher forcing during training. Therefore, the $Z$ vector is feed as the initial hidden state of the decoder. Each time window from the input $\textbf{x}_i+1$ is feed into the same two layer fully connected layer and transformed into a 256 vector. Each window is then passed to a unidirectional GRU layer followed by two fully connected layers with ReLU activation that reconstruct the previous time window $\textbf{x}'_i$.

\newpage

\section{Baseline Methods: Diabetes Dataset}
\label{apd:baseline}

\begin{figure*}[!h]
\centering
{%
\subfigure[Kaplan-Meier curves for PCA k-means clusters (k=3)]{
  \includegraphics[width=40mm]{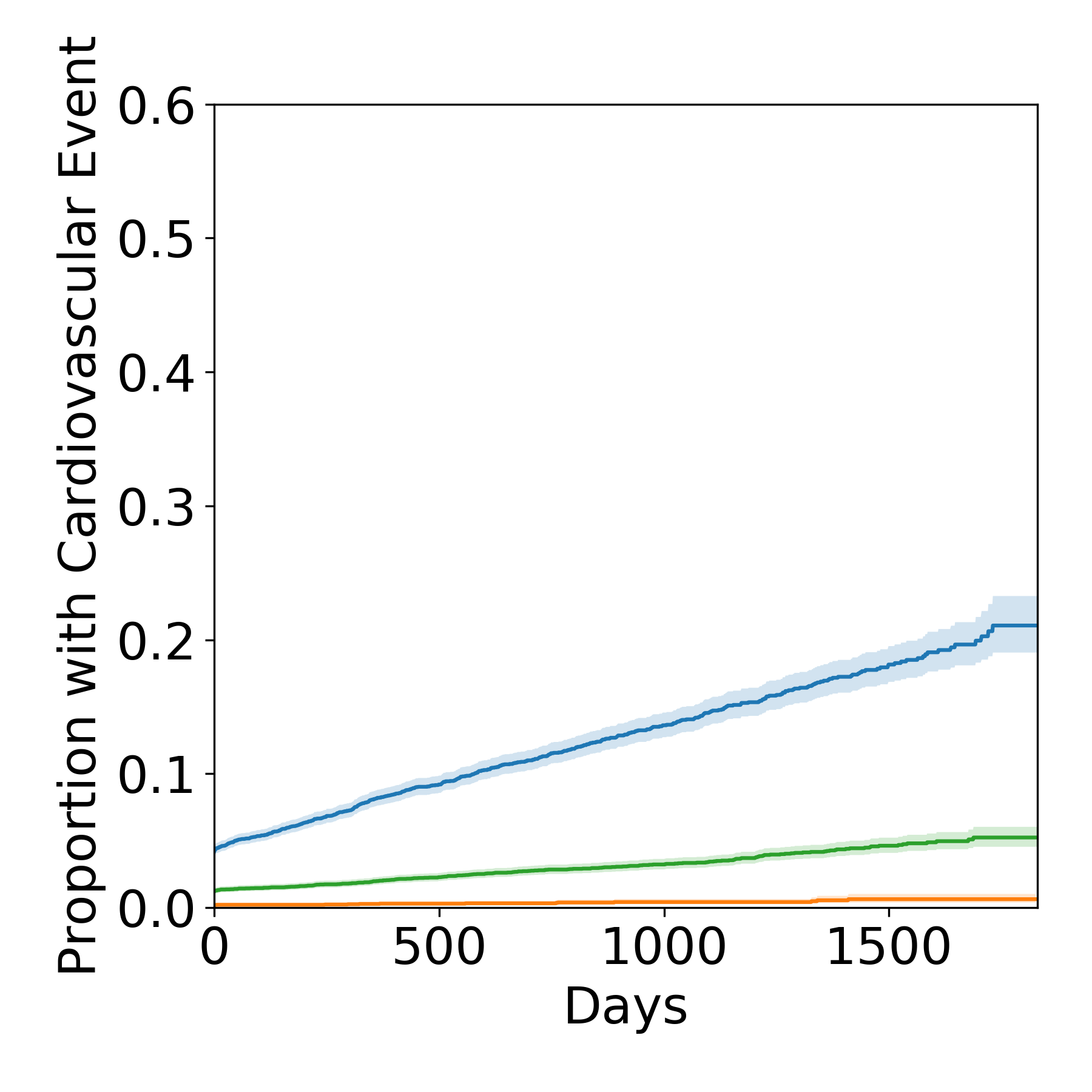}
  \label{fig:baselineA}
}
\subfigure[Kaplan-Meier curves for RSF clusters (k=3)]{
  \includegraphics[width=40mm]{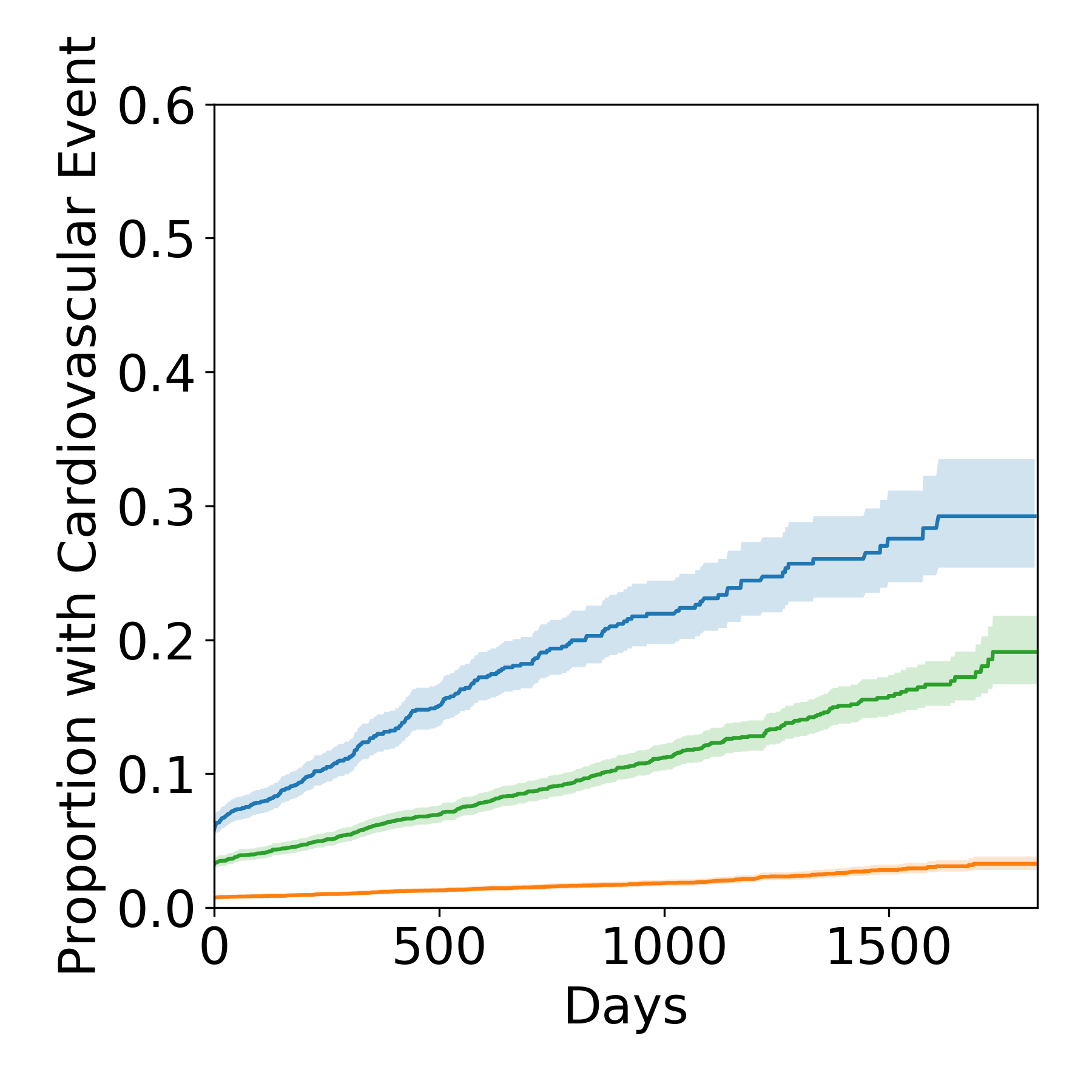}
  \label{fig:baselineB}
} \\
\hspace{0mm}
\subfigure[Kaplan-Meier curves for PCA k-means clusters (k=5)]{
  \includegraphics[width=40mm]{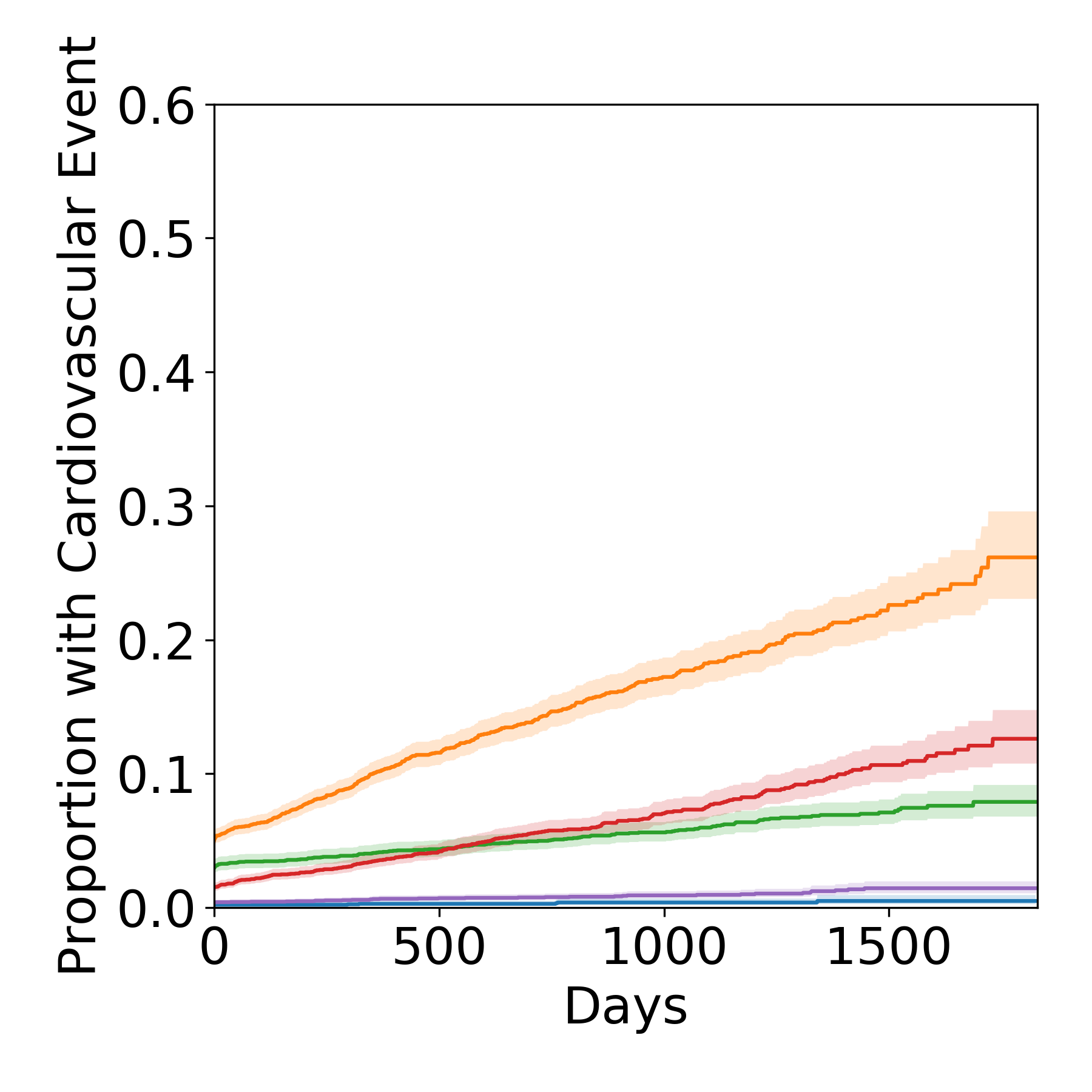}
  \label{fig:baselineC}
}
\subfigure[Kaplan-Meier curves for RSF clusters (k=5)]{
  \includegraphics[width=40mm]{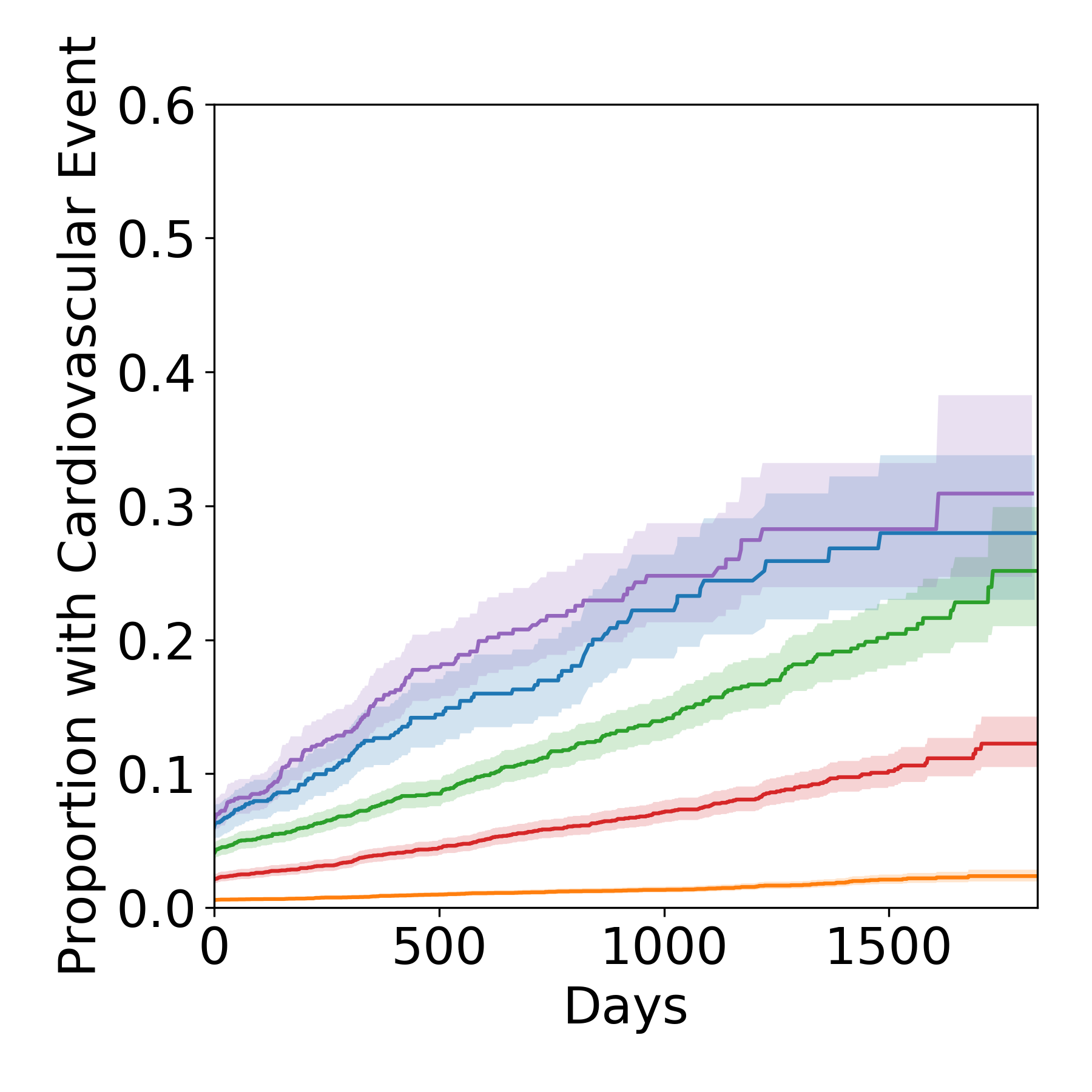}
  \label{fig:baselineD}
}
}
\caption{Kaplan-Meier curves for the clusters of the baseline unsupervised and supervised methods.}
\label{fig:baselineKM}
\end{figure*}

Table \ref{tab:baseline} shows the test statistics of the log rank test between the \gls{km} curves from the clusters obtained from the baseline methods (PCA k-means and Random Survival Forests). The Kaplan-Meier curves using the baseline methods for $k=3$ and $k=5$ are shown in Figure \ref{fig:baselineKM}, which correspond to the results shown in Figure \ref{fig:KMcurves} which use \gls{lps-co}.

\begin{table}[!h]
\centering
\begin{tabular}{lcc}
\hline
Clusters & PCA k-means & \gls{rsf} \\
\hline
2 & $870\pm3$ & $1290\pm52$ \\
3 & $1205\pm4$ & $1553\pm77$ \\
4 & $1358\pm2$ & $1564\pm67$ \\
5 & $1357\pm1$ & $1608\pm52$ \\
6 & $1415\pm5$ & $1544\pm35$ \\
7 & $1431\pm2$ & $1578\pm27$ \\
\hline
\end{tabular}
\caption{\label{tab:baseline}Log rank test statistic between reconstruction PCA k-means and \gls{rsf} clusters, showing separation of outcomes between the discovered clusters for each k.}
\end{table}

\newpage

\section{Additional Cluster Comparison Metrics}
\label{apd:clustermetrics}

Additional clustering metrics can be used to compare the similarity of the clusters discovered using a purely unsupervised, supervised, or combined loss version of \gls{lps-co}. Table \ref{tab:NMI} shows the normalised mutual information scores between the models for different numbers of clusters, showing the same trend as Table \ref{tab:ARI}.

\begin{table}[!h]
\resizebox{\columnwidth}{!}{
\begin{tabular}{lccc}
\hline
Clusters & Recon.-Combined & Outcome-Combined & Recon.-Outcome  \\
\hline
2 & $0.05\pm0.06$ & $0.16\pm0.09$ & $0.06\pm0.09$ \\
3 & $0.26\pm0.18$ & $0.29\pm0.17$ & $0.11\pm0.13$ \\
4 & $0.12\pm0.02$ & $0.24\pm0.09$ & $0.08\pm0.05$ \\
5 & $0.21\pm0.03$ & $0.26\pm0.10$ & $0.10\pm0.05$ \\
6 & $0.18\pm0.08$ & $0.29\pm0.12$ & $0.13\pm0.06$ \\
7 & $0.20\pm0.05$ & $0.35\pm0.07$ & $0.11\pm0.02$ \\
\hline
\end{tabular}
}
\caption{\label{tab:NMI}Normalised mutual information scores between pairs of \gls{lps-co} clusters from different loss weights, showing similarities between the discovered clusters for each k.}
\end{table}

\newpage

\section{Data Summary}
\label{apd:datasummary}

Table \ref{tab:demographics} shows a summary of demographic information for the diabetes cohort, including the distributions of gender, ethnicity, and age across the $29,229$ patients.

Table \ref{tab:lims} shows the distribution ($10$th, $50$th, $90$th percentiles) of the most commonly occurring laboratory values for the diabetes cohort patients. The table is ordered by the total number of patients who have at least one measurement of the laboratory measure across their trajectory (summarised by the Counts per Patient column).

Tables \ref{tab:Pdiagnoses} and \ref{tab:Sdiagnoses} summarise the most occurring $20$ primary and secondary diagnoses codes respectively. The tables are ordered by the total number of patients in the diabetes cohort who have at least one recorded diagnosis of the code across their trajectory. Similarly, Table \ref{tab:procedures} shows the most occurring procedure codes appearing in the diabetes cohort patients, and Table \ref{tab:drugs} shows to most occurring medication codes.

\break
\begin{table}[!h]
\centering
\begin{tabular}{ccc}
\multicolumn{3}{c} {\bf Gender} \\
{\it Male} & {\it Female} & {\it Unknown}  \\
\hline
16,824 & 12,403 & 2\\
\hline
\\ \multicolumn{3}{c} {\bf Ethnicity} \\
\hline
{\it White British} & {\it Not Stated} & {\it Other}  \\
\hline
20,271 & 5,256  & 3,702 \\
\hline
\\ \multicolumn{3}{c} {\bf Age} \\
\hline
{\it 10th} & {\it 50th} & {\it 90th}  \\
\hline
46.84  & 69.04 & 84.82\\
\hline
\end{tabular}
\caption{\label{tab:demographics} Summary of demographic information of the diabetes cohort. Counts of gender, ethnicity, and percentiles of age are shown.}
\end{table}

\begin{table*}[!ht]
\resizebox{\textwidth}{!}{
\centering
\begin{tabular}{|l|c|ccc|}
\hline
 & Counts per Patient & \multicolumn{3}{c}{Percentile} \\
Laboratory Measurement Name & (Total=29,229) & 10th & 50th & 90th \\
\hline
Blood Creatinine (umol/l) & 26,769 & 54.00 & 78.00 & 142.00\\
Blood Sodium (mmol/l) & 26,746 & 134.50 & 138.75 & 141.75\\
Blood Potassium (mmol/l) & 26,740 & 3.70 & 4.10 & 4.70\\
Blood Estimated Glomerular Filtration Rate eGFR (ml/min/1.73m2) & 26,669 & 38.00 & 78.00 & 90.00\\
Blood White Blood Cells WBC (10e9/l) & 26,196 & 5.62 & 8.10 & 11.81\\
Blood Haemoglobin (g/dl) & 26,195 & 10.28 & 13.00 & 15.20\\
Blood Mean Corpuscular Haemoglobin Concentration MCHC (g/l) & 26,193 & 311.00 & 327.00 & 342.00\\
Blood Mean Corpuscular Volume MCV (fl) & 26,193 & 82.95 & 89.80 & 96.70\\
Blood Mean Corpuscular Haemoglobin MCH (pg) & 26,193 & 26.60 & 29.50 & 31.90\\
Blood Red Blood Cell RBC Count (10e12/l) & 26,193 & 3.54 & 4.44 & 5.18\\
Blood Haematocrit HCT (l/l) & 26,193 & 0.32 & 0.40 & 0.46\\
Blood Platelets (10e9/l) & 26,188 & 170.00 & 251.00 & 361.00\\
Blood Albumin (g/l) & 24,998 & 29.00 & 37.00 & 42.00\\
Alkaline Phosphatase ALP (iu/l) & 24,956 & 57.00 & 88.50 & 187.50\\
Blood Bilirubin (umol/l) & 24,880 & 5.00 & 9.00 & 17.00\\
Alanine Aminotransferase ALT (iu/l) & 24,879 & 12.00 & 21.00 & 45.00\\
Blood Urea (mmol/l) & 23,702 & 3.75 & 6.00 & 12.80\\
Blood Mean Platelet Volume MPV (fl) & 22,331 & 9.50 & 10.60 & 12.00\\
Blood C Reactive Protein CRP (mg/l) & 20,866 & 1.10 & 10.45 & 100.00\\
Blood HDL Cholesterol (mmol/l) & 18,661 & 0.80 & 1.10 & 1.65\\
Blood Total Cholesterol (mmol/l) & 18,661 & 3.10 & 4.15 & 5.75\\
Blood Cholesterol HDL Ratio (ratio) & 18,656 & 2.50 & 3.65 & 5.50\\
Blood Glucose (mmol/l) & 16,917 & 5.50 & 8.20 & 14.65\\
Thyroid Stimulating Hormone TSH (mu/l) & 16,198 & 0.68 & 1.67 & 3.60\\
Blood International Normalised Ratio INR (ratio) & 15,098 & 1.00 & 1.05 & 2.20\\
Urine Creatinine (mmol/24h) & 14,716 & 3.41 & 7.45 & 14.70\\
Blood Triglycerides (mmol/l) & 14,096 & 0.84 & 1.57 & 3.14\\
Urine Albumin (mg/l) & 13,768 & 0.01 & 0.01 & 0.12\\
Blood LDL Cholesterol (mmol/l) & 13,525 & 1.30 & 2.15 & 3.50\\
Urine Albumin Creatinine Ratio (mg/mmol) & 11,557 & 0.60 & 2.00 & 20.90\\
Blood B12 (pg/ml) & 9,146 & 202.25 & 354.00 & 751.00\\
Blood Ferritin (ug/l) & 8,906 & 17.80 & 88.91 & 413.23\\
Blood Folate (ug/l) & 8,044 & 3.20 & 6.40 & 14.30\\
Blood Iron (umol/l) & 7,448 & 5.00 & 11.10 & 19.31\\
Blood Transferrin (g/l) & 7,448 & 1.81 & 2.58 & 3.36\\
Blood Transferrin Saturation (
Blood Troponin I (ng/l) & 6,126 & 20.00 & 40.00 & 897.25\\
Blood Erythrocyte Sedimentation Rate ESR (mm/h) & 4,538 & 2.00 & 14.00 & 53.00\\
Blood Vitamin D VitD (nmol/l) & 3,554 & 18.00 & 41.00 & 79.07\\
Blood Gamma Glutamyl Transferase GGT (iu/l) & 3,427 & 18.00 & 47.25 & 262.25\\
Blood Thyroxine T4 (pmol/l) & 3,278 & 11.08 & 14.10 & 18.90\\
\hline
\end{tabular}
}
\caption{\label{tab:lims} Values of most frequent laboratories values.  The counts correspond to the number of patients that have at least one measurement along its trajectory. The percentiles presented correspond to the distribution of the median values of each patient along its trajectory.}
\end{table*}

\begin{table*}[!ht]
\centering
{
\begin{tabular}{|l|c|}
\hline
 & Counts per Patient \\
Primary Diagnoses &  (Total=29,229)\\
\hline
H26.9 Cataract, unspecified & 1,595 \\
I25.1 Atherosclerotic heart disease & 1,449 \\
R07.4 Chest pain, unspecified & 1,286 \\
N39.0 Urinary tract infection, site not specified & 712 \\
K63.5 Polyp of colon & 710 \\
J18.9 Pneumonia, unspecified & 573 \\
N18.5 Chronic kidney disease, stage 5 & 558 \\
J18.1 Lobar pneumonia, unspecified & 554 \\
D63.8 Anaemia in other chronic diseases classified elsewhere & 505 \\
D50.9 Iron deficiency anaemia, unspecified & 450 \\
J22 Unspecified acute lower respiratory infection & 426 \\
Z03.5 Observation for other suspected cardiovascular diseases & 415 \\
G47.3 Sleep apnoea & 412 \\
H25.1 Senile nuclear cataract & 403 \\
A09.9 Gastroenteritis and colitis of unspecified origin & 383 \\
H36.0 Diabetic retinopathy & 382 \\
N17.9 Acute renal failure, unspecified & 367 \\
I21.4 Acute subendocardial myocardial infarction & 359 \\
R06.0 Dyspnoea & 356 \\
C44.3 Skin of other and unspecified parts of face & 347 \\
\hline
\end{tabular}
}
\caption{\label{tab:Pdiagnoses}Counts primary diagnosis codes which occur at least once in the trajectory of patients in the diabetes cohort.}

\centering
{
\begin{tabular}{|l|c|}
\hline
 & Counts per Patient \\
Secondary Diagnoses &  (Total=29,229)\\
\hline
E11.9 Non-insulin-dependent diabetes mellitus: Without complications & 24,425 \\
I10 Essential (primary) hypertension & 17,190 \\
Z92.2 Personal history of long-term (current) use of other medicaments & 6,969 \\
I25.9 Chronic ischaemic heart disease, unspecified & 3,267 \\
Z92.1 Personal history of long-term (current) use of anticoagulants & 3,196 \\
Z86.7 Personal history of diseases of the circulatory system & 2,995 \\
E78.0 Pure hypercholesterolaemia & 2,980 \\
I25.1 Atherosclerotic heart disease & 2,951 \\
Z86.4 Personal history of psychoactive substance abuse & 2,892 \\
J45.9 Asthma, unspecified & 2,887 \\
F17.1 Harmful use & 2,881 \\
I25.2 Old myocardial infarction & 2,702 \\
Z88.0 Personal history of allergy to penicillin & 2,531 \\
E10.9 Insulin-dependent diabetes mellitus: Without complications & 2,008 \\
I48 Atrial fibrillation and flutter & 1,930 \\
F32.9 Depressive episode, unspecified & 1,915 \\
I48.9 Atrial fibrillation and atrial flutter, unspecified & 1,898 \\
N17.9 Acute renal failure, unspecified & 1,881 \\
J44.9 Chronic obstructive pulmonary disease, unspecified & 1,843 \\
E03.9 Hypothyroidism, unspecified & 1,826 \\
\hline
\end{tabular}
}
\caption{\label{tab:Sdiagnoses}Counts secondary diagnosis codes which occur at least once in the trajectory of patients in the diabetes cohort.}
\end{table*}

\begin{table*}[!h]
\centering
{
\begin{tabular}{|l|c|}
\hline
 & Counts per Patient \\
Procedures &  (Total=29,229)\\
\hline
Y98.1 Radiology of one body area (or < 20 minutes) & 5,471 \\
Z94.2 Right sided operation & 5,083 \\
Z94.3 Left sided operation & 4,860 \\
Y53.4 Approach to organ under fluoroscopic control & 3,421 \\
U20.1 Transthoracic echocardiography & 3,335 \\
Y97.3 Radiology with post contrast & 2,917 \\
U21.2 Computed tomography NEC & 2,853 \\
U05.1 Computed tomography of head & 2,776 \\
Z94.1 Bilateral operation & 2,106 \\
G45.1 Fibreoptic endoscopic examination of upper gastrointestinal tract 
& 1,985 \\
Z92.6 Abdomen NEC & 1,815 \\
Z27.4 Duodenum & 1,798 \\
O16.1 Pelvis NEC & 1,788 \\
C75.1 Insertion of prosthetic replacement for lens NEC & 1,779 \\
C87.3 Tomography evaluation of retina & 1,777 \\
C71.2 Phacoemulsification of lens & 1,766 \\
Y98.2 Radiology of two body areas & 1,620 \\
Y53.2 Approach to organ under ultrasonic control & 1,555 \\
U10.6 Myocardial perfusion scan & 1,353 \\
Z28.6 Sigmoid colon & 1,068 \\
\hline
\end{tabular}
}
\caption{\label{tab:procedures}Counts procedure codes which occur at least once in the trajectory of patients in the diabetes cohort.}

\centering
{
\begin{tabular}{|l|c|}
\hline
 & Counts per Patient \\
Medications &  (Total=29,229)\\
\hline
Analgesics (INP) & 14,451 \\
Anticoagulants And Protamine (INP) & 12,385 \\
Analgesics (TTA) & 11,718 \\
Antibacterial Drugs (INP) & 11,686 \\
Drugs Used In Diabetes (TTA) & 11,558 \\
Lipid-Regulating Drugs (TTA) & 10,928 \\
Drugs Used In Diabetes (INP) & 10,708 \\
Hypertension and Heart Failure (TTA) & 9,938 \\
Lipid-Regulating Drugs (INP) & 8,666 \\
Antisecretory Drugs+Mucosal Protectants (TTA) & 8,563 \\
Antisecretory Drugs+Mucosal Protectants (INP) & 8,047 \\
Hypertension and Heart Failure (INP) & 7,610 \\
Antiplatelet Drugs (TTA) & 7,022 \\
Antibacterial Drugs (TTA) & 6,791 \\
Acute Diarrhoea (INP) & 6,688 \\
Antiplatelet Drugs (INP) & 6,362 \\
Acute Diarrhoea (TTA) & 6,124 \\
Drugs Used In Nausea And Vertigo (INP) & 5,999 \\
Beta-Adrenoceptor Blocking Drugs (TTA) & 5,885 \\
Diuretics (TTA) & 5,586 \\
\hline
\end{tabular}
}
\caption{\label{tab:drugs}Counts medication codes which occur at least once in the trajectory of patients in the diabetes cohort.}
\end{table*}

\end{document}